\documentclass{article} %
\usepackage{colm2024_conference}

\usepackage{booktabs}
\usepackage{graphicx}
\usepackage{enumitem}
\usepackage{wrapfig}
\usepackage{algorithm}
\usepackage{algpseudocode}

\usepackage{microtype}
\usepackage{amsmath}
\usepackage{colortbl}
\usepackage[utf8]{inputenc}
\definecolor{lightgray}{rgb}{0.9,0.9,0.9}
\usepackage{caption}
\usepackage{subcaption}
\usepackage{xcolor}
\usepackage{setspace}
\usepackage{url}
\usepackage{multirow}
\usepackage{colortbl}
\usepackage{tabularx}
\usepackage{blindtext}
\usepackage{pgfplots}
\pgfplotsset{compat=1.18} 
\usepackage{tikz}
\usetikzlibrary{er,positioning,bayesnet}
\usepackage{makecell}
\usepackage{tipa}
\usepackage{siunitx}
\usepackage{nicefrac}
\usepackage{tocloft}
\usepackage{listings}
\usepackage[raster,skins]{tcolorbox} %
\usepackage{xltabular}
\usepackage{adjustbox}
\usepackage{xurl}

\usepackage{amsmath,amsfonts,bm}

\def\eqref#1{equation~\ref{#1}}

\def\1{\bm{1}}

\DeclareMathAlphabet{\mathsfit}{\encodingdefault}{\sfdefault}{m}{sl}
\SetMathAlphabet{\mathsfit}{bold}{\encodingdefault}{\sfdefault}{bx}{n}

\title{Qwen2.5 Technical Report}

\author{
\bf Qwen Team
}

\begin{document}

\maketitle

\begin{abstract}

In this report, we introduce Qwen2.5, a comprehensive series of large language models (LLMs) designed to meet diverse needs. 
Compared to previous iterations, Qwen 2.5 has been significantly improved during both the pre-training and post-training stages. 
In terms of pre-training, we have scaled the high-quality pre-training datasets from the previous 7 trillion tokens to 18 trillion tokens.
This provides a strong foundation for %
common sense, expert knowledge, and reasoning capabilities.
In terms of post-training, we implement intricate supervised finetuning with over 1 million samples, as well as multistage reinforcement learning, including offline learning DPO and online learning GRPO. 
Post-training techniques significantly enhance human preference, and notably improve long text generation, structural data analysis, and instruction following.

\smallskip
To handle diverse and varied use cases effectively, 
we present Qwen2.5 LLM series in rich configurations.
The open-weight offerings include base models and instruction-tuned models in sizes of 0.5B, 1.5B, 3B, 7B, 14B, 32B, and 72B parameters. Quantized versions of the instruction-tuned models are also provided. Over 100 models can be accessed from Hugging Face Hub, ModelScope, and Kaggle.
In addition, for hosted solutions, the proprietary models currently include two mixture-of-experts (MoE) variants: Qwen2.5-Turbo and Qwen2.5-Plus, both available from \href{https://www.alibabacloud.com/en/product/modelstudio}{Alibaba Cloud Model Studio}.

Qwen2.5 has demonstrated top-tier performance on a wide range of benchmarks evaluating language understanding, reasoning, mathematics, coding, human preference alignment, etc. Specifically, the open-weight flagship Qwen2.5-72B-Instruct outperforms a number of open and proprietary models and demonstrates competitive performance to the state-of-the-art open-weight model, Llama-3-405B-Instruct, which is around 5 times larger.
Qwen2.5-Turbo and Qwen2.5-Plus offer superior cost-effectiveness while performing competitively against GPT-4o-mini and GPT-4o respectively.
Additionally, as the foundation, Qwen2.5 models have been instrumental in training specialized models such as Qwen2.5-Math~\citep{qwen2.5math}, Qwen2.5-Coder~\citep{qwen2.5coder}, QwQ~\citep{qwq}, and multimodal models.

\end{abstract}

\vfill

\begin{figure}[hbp]
    \centering
    \includegraphics[width=0.92\textwidth]{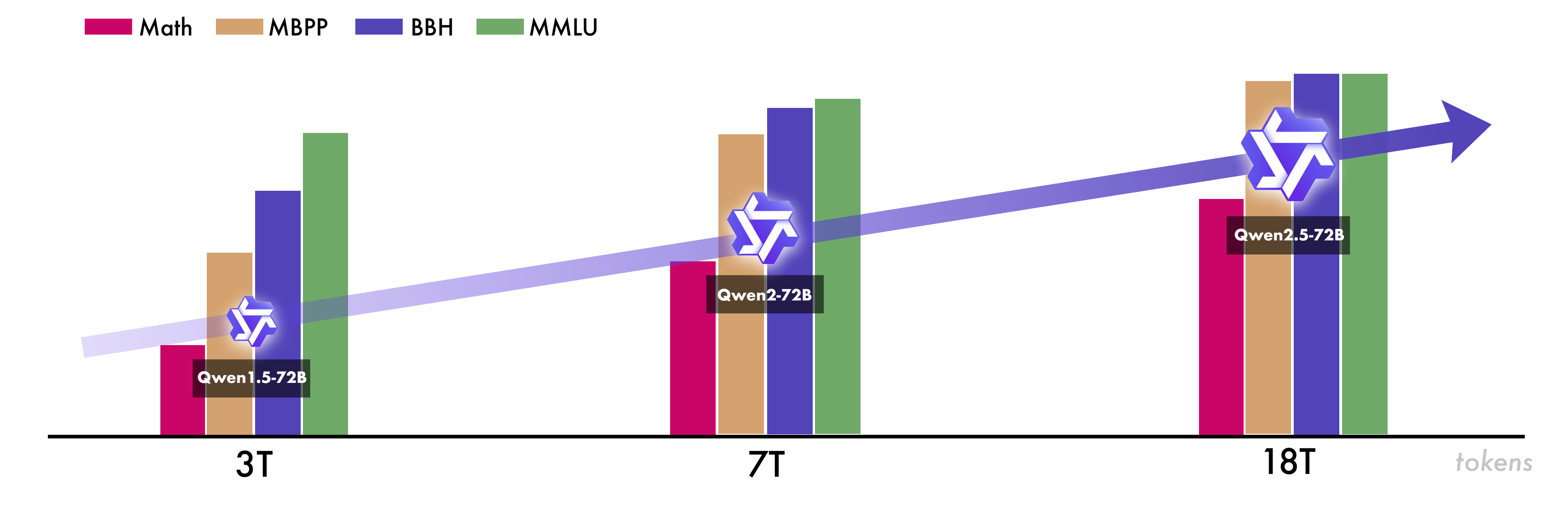}
    \caption{In the iterative development of the Qwen series, data scaling has played a crucial role. Qwen~2.5, which leverages 18 trillion tokens for pre-training, has demonstrated the most advanced capabilities within the Qwen series, especially in terms of domain expertise, underscoring the importance of scale together with mixture in enhancing the model's capabilities.}
    \label{fig:intro}
\end{figure}

\vfill

\newpage

\section{Introduction}
\label{sec:intro}

The sparks of artificial general intelligence (AGI) are increasingly visible through the fast development of large foundation models, notably large language models (LLMs)~\citep{gpt3,gpt4,gpt4o,gemini,claude,claude2,claude3,qwen,qwen2,llama,llama2,llama3}. The continuous advancement in model and data scaling, combined with the paradigm of large-scale pre-training followed by high-quality supervised fine-tuning (SFT) and reinforcement learning from human feedback (RLHF)~\citep{instructgpt}, has enabled large language models (LLMs) to develop emergent capabilities in language understanding, generation, and reasoning. Building on this foundation, recent breakthroughs in inference time scaling, particularly demonstrated by o1~\citep{o1}, have enhanced LLMs' capacity for deep thinking through step-by-step reasoning and reflection. These developments have elevated the potential of language models, suggesting they may achieve significant breakthroughs in scientific exploration as they continue to demonstrate emergent capabilities indicative of more general artificial intelligence.

Besides the fast development of model capabilities, the recent two years have witnessed a burst of open (open-weight) large language models in the LLM community, for example, the Llama series~\citep{llama, llama2, llama3}, Mistral series~\citep{mistral, mixtral}, and our Qwen series~\citep{qwen, qwen2, codeqwen, qwen2.5coder, qwen2math, qwen2.5math}. The open-weight models have democratized the access of large language models to common users and developers, enabling broader research participation, fostering innovation through community collaboration, and accelerating the development of AI applications across diverse domains.

Recently, we release the details of our latest version of the Qwen series, Qwen2.5. In terms of the open-weight part, we release pre-trained and instruction-tuned models of 7 sizes, including 0.5B, 1.5B, 3B, 7B, 14B, 32B, and 72B, and we provide not only the original models in bfloat16 precision but also the quantized models in different precisions. Specifically, the flagship model Qwen2.5-72B-Instruct demonstrates competitive performance against the state-of-the-art open-weight model, Llama-3-405B-Instruct, which is around 5 times larger. Additionally, we also release the proprietary models of Mixture-of-Experts (MoE, \citealp{gshard, switch_transformer, stmoe}), namely Qwen2.5-Turbo and Qwen2.5-Plus\footnote{Qwen2.5-Turbo is identified as \texttt{qwen-turbo-2024-11-01} and Qwen2.5-Plus is identified as \texttt{qwen-plus-2024-xx-xx} (to be released) in the API.}, which performs competitively against GPT-4o-mini and GPT-4o respectively.

In this technical report, we introduce Qwen2.5, the result of our continuous endeavor to create better LLMs. Below, we show the key features of the latest version of Qwen:

\begin{itemize}
    \item \textbf{Better in Size}: Compared with Qwen2, in addition to 0.5B, 1.5B, 7B, and 72B models, Qwen2.5 brings back the 3B, 14B, and 32B models, which are more cost-effective for resource-limited scenarios and are under-represented in the current field of open foundation models. Qwen2.5-Turbo and Qwen2.5-Plus offer a great balance among accuracy, latency, and cost.
    
    \item \textbf{Better in Data}: The pre-training and post-training data have been improved significantly.
    The pre-training data increased from 7 trillion tokens to 18 trillion tokens, with focus on knowledge, coding, and mathematics.
    The pre-training is staged to allow transitions among different mixtures.
    The post-training data amounts to 1 million examples, across the stage of supervised finetuning (SFT, \citealp{instructgpt}), direct preference optimization (DPO, \citealp{rafailov2024direct}), and group relative policy optimization (GRPO, ~\citealp{deepseekmath}).
    
    \item \textbf{Better in Use}: Several key limitations of Qwen2 in use have been eliminated, including larger generation length (from 2K tokens to 8K tokens), better support for structured input and output, (e.g., tables and JSON), and easier tool use. In addition, Qwen2.5-Turbo supports a context length of up to 1 million tokens.
\end{itemize}

\section{Architecture \& Tokenizer}

Basically, the Qwen2.5 series include dense models for opensource, namely Qwen2.5-0.5B / 1.5B / 3B / 7B / 14B / 32B / 72B, and MoE models for API service, namely Qwen2.5-Turbo and Qwen2.5-Plus. Below, we provide details about the architecture of models.

For dense models, we maintain the Transformer-based decoder architecture~\citep{transformer,gpt} as Qwen2~\citep{qwen2}. The architecture incorporates several key components: Grouped Query Attention (GQA, \citealp{gqa}) for efficient KV cache utilization, SwiGLU activation function~\citep{glu} for non-linear activation, Rotary Positional Embeddings (RoPE, \citealp{rope}) for encoding position information, QKV bias~\citep{qkv_bias} in the attention mechanism and RMSNorm~\citep{rmsnorm} with pre-normalization to ensure stable training.

Building upon the dense model architectures, we extend it to MoE model architectures. This is achieved by replacing standard feed-forward network (FFN) layers with specialized MoE layers, where each layer comprises multiple FFN experts and a routing mechanism that dispatches tokens to the top-K experts. Following the approaches demonstrated in Qwen1.5-MoE~\citep{qwen2}, we implement fine-grained expert segmentation~\citep{deepseekmoe} and shared experts routing~\citep{deepspeedmoe, deepseekmoe}. These architectural innovations have yielded substantial improvements in model performance across downstream tasks.

For tokenization, we utilize Qwen's tokenizer~\citep{qwen}, which implements byte-level byte-pair encoding (BBPE,~\citealp{gpt3,wang2020neural,sennirch2016neural}) with a vocabulary of 151,643 regular tokens. We have expanded the set of control tokens from 3 to 22 compared to previous Qwen versions, adding two new tokens for tool functionality and allocating the remainder for other model capabilities. This expansion establishes a unified vocabulary across all Qwen2.5 models, enhancing consistency and reducing potential compatibility issues.

\begin{table}[tbp]
\caption{Model architecture and license of Qwen2.5 open-weight models.}
\small
\centering
\begin{tabular}{@{}lcccccccc@{}} 
\toprule
Models  & Layers & Heads (Q / KV) & Tie Embedding & Context / Generation Length  & License \\
\midrule
0.5B  & 24 & 14 / 2 & Yes & 32K / 8K & Apache 2.0 \\
1.5B  & 28 & 12 / 2 & Yes & 32K / 8K & Apache 2.0 \\
3B  & 36 & 16 / 2 & Yes & 32K / 8K & Qwen Research \\
7B  & 28 & 28 / 4 & No & 128K / 8K & Apache 2.0 \\
14B & 48 & 40 / 8 & No & 128K / 8K & Apache 2.0 \\
32B & 64 & 40 / 8 & No & 128K / 8K & Apache 2.0 \\
72B  & 80 & 64 / 8 & No & 128K / 8K & Qwen \\
\bottomrule
\end{tabular}
\end{table}

\section{Pre-training}

Our language model pre-training process consists of several key components. First, we carefully curate high-quality training data through sophisticated filtering and scoring mechanisms, combined with strategic data mixture. Second, we conduct extensive research on hyperparameter optimization to effectively train models at various scales. Finally, we incorporate specialized long-context pre-training to enhance the model's ability to process and understand extended sequences. Below, we detail our approaches to data preparation, hyperparameter selection, and long-context training.

\label{sec:pre}

\subsection{Pre-training Data}

Qwen2.5 demonstrates significant enhancements in pre-training data quality compared to its predecessor Qwen2. These improvements stem from several key aspects:

\begin{enumerate}[label=(\arabic*)]
    \item \textbf{Better data filtering}. High-quality pre-training data is crucial for model performance, making data quality assessment and filtering a critical component of our pipeline. We leverage Qwen2-Instruct models as data quality filters that perform comprehensive, multi-dimensional analysis to evaluate and score training samples. The filtering method represents a significant advancement over our previous approach used for Qwen2, as it benefits from Qwen2's expanded pre-training on a larger multilingual corpus. The enhanced capabilities enable more nuanced quality assessment, resulting in both improved retention of high-quality training data and more effective filtering of low-quality samples across multiple languages.

    \item \textbf{Better math and code data}. During the pre-training phase of Qwen2.5, we incorporate training data from Qwen2.5-Math~\citep{qwen2.5math} and Qwen2.5-Coder~\citep{qwen2.5coder}. This data integration strategy proves highly effective, as these specialized datasets are instrumental in achieving state-of-the-art performance on mathematical and coding tasks. By leveraging these high-quality domain-specific datasets during pre-training, Qwen2.5 inherits strong capabilities in both mathematical reasoning and code generation.

    \item \textbf{Better synthetic data}. To generate high-quality synthetic data, particularly in mathematics, code, and knowledge domains, we leverage both Qwen2-72B-Instruct~\citep{qwen2} and Qwen2-Math-72B-Instruct~\citep{qwen2math}. The quality of this synthesized data is further enhanced through rigorous filtering using our proprietary general reward model and the specialized Qwen2-Math-RM-72B~\citep{qwen2math} model.

    \item \textbf{Better data mixture}. To optimize the pre-training data distribution, we employ Qwen2-Instruct models to classify and balance content across different domains. Our analysis revealed that domains like e-commerce, social media, and entertainment are significantly overrepresented in web-scale data, often containing repetitive, template-based, or machine-generated content. Conversely, domains such as technology, science, and  academic research, while containing higher-quality information, are traditionally underrepresented. Through strategic down-sampling of overrepresented domains and up-sampling of high-value domains, we ensure a more balanced and information-rich training dataset that better serves our model's learning objectives.
\end{enumerate}

Building on these techniques, we have developed a larger and higher-quality pre-training dataset, expanding from the 7 trillion tokens used in Qwen2~\citep{qwen2} to \textbf{18 trillion} tokens.

\subsection{Scaling Law for Hyper-parameters}
We develop scaling laws for hyper-parameter based on the pre-training data of Qwen2.5~\citep{chinchilla,kaplanscaling}. While previous studies~\citep{llama3,falcon,chinchilla} primarily used scaling laws to determine optimal model sizes given compute budgets, we leverage them to identify optimal hyperparameters across model architectures. Specifically, our scaling laws help determine key training parameters like batch size $B$ and learning rate $\mu$ for both dense models and MoE models of varying sizes.

Through extensive experimentation, we systematically study the relationship between model architecture and optimal training hyper-parameters. Specifically, we analyze how the optimal learning rate $\mu_\textrm{opt}$ and batch size $B_\textrm{opt}$ vary with model size $N$ and pre-training data size $D$. Our experiments cover a comprehensive range of architectures, including dense models with 44M to 14B parameters and MoE models with 44M to 1B activated parameters, trained on datasets ranging from 0.8B to 600B tokens. Using these optimal hyper-parameter predictions, we then model the final loss as a function of model architecture and training data scale.

Additionally, we leverage scaling laws to predict and compare the performance of MoE models with varying parameter counts against their dense counterparts. This analysis guides our hyper-parameter configuration for MoE models, enabling us to achieve performance parity with specific dense model variants (such as Qwen2.5-72B and Qwen2.5-14B) through careful tuning of both activated and total parameters.

\subsection{Long-context Pre-training}

For optimal training efficiency, Qwen2.5 employs a two-phase pre-training approach: an initial phase with a 4,096-token context length, followed by an extension phase for longer sequences. Following the strategy used in Qwen2, we extend the context length from 4,096 to 32,768 tokens during the final pre-training stage for all model variants except Qwen2.5-Turbo. Concurrently, we increase the base frequency of RoPEfrom 10,000 to 1,000,000 using the ABF technique~\citep{ropeabf}.

For Qwen2.5-Turbo, we implement a progressive context length expansion strategy during training, advancing through four stages: 32,768 tokens, 65,536 tokens, 131,072 tokens, and ultimately 262,144 tokens, with a RoPE base frequency of 10,000,000. At each stage, we carefully curate the training data to include 40\% sequences at the current maximum length and 60\% shorter sequences. This progressive training methodology enables smooth adaptation to increasing context lengths while maintaining the model's ability to effectively process and generalize across sequences of varying lengths.

To enhance our models' ability to process longer sequences during inference, we implement two key strategies: YARN~\citep{yarn} and Dual Chunk Attention~(DCA, \citealp{chunkllama}). Through these innovations, we achieve a four-fold increase in sequence length capacity, enabling Qwen2.5-Turbo to handle up to \textbf{1 million} tokens and other models to process up to 131,072 tokens. Notably, these approaches not only improve the modeling of long sequences by reducing perplexity but also maintain the models' strong performance on shorter sequences, ensuring consistent quality across varying input lengths.

\section{Post-training}

\label{sec:post}

Qwen 2.5 introduces two significant advancements in its post-training design compared to Qwen 2:

\begin{enumerate}[label=(\arabic*)]

    \item \textbf{Expanded Supervised Fine-tuning Data Coverage:} The supervised fine-tuning process leverages a massive dataset comprising millions of high-quality examples.
    This expansion specifically addresses key areas where the previous model showed limitations, such as long-sequence generation, mathematical problem-solving, coding, instruction-following, structured data understanding, logical reasoning, cross-lingual transfer, and robust system instruction.

    \item \textbf{Two-stage Reinforcement Learning:} The reinforcement learning (RL) process in Qwen 2.5 is divided into two distinct stages: Offline RL and Online RL. 

        \begin{itemize}

            \item \textit{Offline RL:} This stage focuses on developing capabilities that are challenging for the reward model to evaluate, such as reasoning, factuality, and instruction-following.
            Through meticulous construction and validation of training data, we ensure that the Offline RL signals are both learnable and reliable~\citep{xiang2024aligning}, enabling the model to acquire those complex skills effectively.
            
            \item \textit{Online RL:} The Online RL phase leverages the reward model's ability to detect nuances in output quality, including truthfulness, helpfulness, conciseness, relevance, harmlessness and debiasing. 
            It enables the model to generate responses that are precise, coherent, and well-structured while maintaining safety and readability. As a result, the model's outputs consistently meet human quality standards and expectations.

        \end{itemize}

\end{enumerate}

\subsection{Supervised Fine-tuning}

In this section, we detail the key enhancements made during the SFT phase of Qwen2.5, focusing on several critical areas:

\begin{enumerate}[label=(\arabic*)]

    \item \textbf{Long-sequence Generation:} 
    Qwen2.5 is capable of generating high-quality content with an output context length of up to 8,192 tokens, a significant advancement over the typical post-training response length, which often remains under 2,000 tokens. To address this gap, we develop long-response datasets~\citep{quan2024language}. We employ back-translation techniques to generate queries for long-text data from pre-training corpora, impose output length constraints, and use Qwen2 to filter out low-quality paired data.

    \item \textbf{Mathematics:} 
    We introduce the chain-of-thought data of Qwen2.5-Math~\citep{qwen2.5math}, which encompasses a diverse range of query sources, including public datasets, K-12 problem collections, and synthetic problems. To ensure high-quality reasoning, we employ rejection sampling~\citep{yuan2023scaling} along with reward modeling and annotated answers for guidance, producing step-by-step reasoning process.

    \item \textbf{Coding:} 
    To enhance coding capabilities, we incorporate the instruction tuning data of Qwen2.5-Coder~\citep{qwen2.5coder}. We use multiple language-specific agents into a collaborative framework, generating diverse and high-quality instruction pairs across nearly 40 programming languages. 
    We expand our instruction dataset by synthesizing new examples from code-related Q\&A websites and gathering algorithmic code snippets from GitHub. A comprehensive multilingual sandbox is used to perform static code checking and validate code snippets through automated unit testing, ensuring code quality and correctness~\citep{dou2024multi,yang2024evaluating}.
    
    \item \textbf{Instruction-following: } 
    To ensure high-quality instruction-following data, we implement a rigorous code-based validation framework. In this approach, LLMs generate both instructions and corresponding verification code, along with comprehensive unit tests for cross-validation. Through execution feedback-based rejection sampling, we carefully curate the training data used for Supervised Fine-Tuning, thereby guaranteeing the model's faithful adherence to intended instructions~\citep{dong2024autoif}.
    
    \item \textbf{Structured Data Understanding: } 
    We develop a comprehensive structured understanding dataset that encompasses both traditional tasks, such as tabular question-answering, fact verification, error correction, and structural understanding, as well as complex tasks involving structured and semi-structured data. By incorporating reasoning chains into the model's responses, we significantly enhance its ability to infer information from structured data, thereby improving its performance across these diverse tasks. This approach not only broadens the scope of the dataset but also deepens the model's capacity to reason and derive meaningful insights from complex data structures.

    \item \textbf{Logical Reasoning:} To enhance the model's logical reasoning capabilities, we introduce a diverse set of 70,000 new queries spanning various domains. These queries encompass multiple-choice questions, true / false questions, and open-ended questions. 
    The model is trained to approach problems systematically, employing a range of reasoning methods such as deductive reasoning, inductive generalization, analogical reasoning, causal reasoning, and statistical reasoning. Through iterative refinement, we systematically filter out data containing incorrect answers or flawed reasoning processes. This process progressively strengthens the model's ability to reason logically and accurately, ensuring robust performance across different types of reasoning tasks.
    
    \item \textbf{Cross-Lingual Transfer:} To facilitate the transfer of the model's general capabilities across languages, we employ a translation model to convert instructions from high-resource languages into various low-resource languages, thereby generating corresponding response candidates. To ensure the accuracy and consistency of these responses, we evaluate the semantic alignment between each multilingual response and its original counterpart. This process preserves the logical structure and stylistic nuances of the original responses, thereby maintaining their integrity and coherence across different languages.
    
    \item \textbf{Robust System Instruction:}  We construct hundreds of general system prompts to improve the diversity of system prompts in post-training, ensuring consistency between system prompts and conversations. Evaluations with different system prompts show that the model maintains good performance~\citep{lu2024large} and reduced variance, indicating improved robustness.
    
    \item \textbf{Response Filtering:}  To evaluate the quality of responses, we employ multiple automatic annotation methods, including a dedicated critic model and a multi-agent collaborative scoring system. Responses are subjected to rigorous assessment, and only those deem flawless by all scoring systems are retained. This comprehensive approach ensures that our outputs maintain the highest quality standards.
    
\end{enumerate}

Ultimately, we construct a dataset of over 1 million SFT examples. The model is fine-tuned for two epochs with a sequence length of 32,768 tokens. To optimize learning, the learning rate is gradually decreased from $7 \times 10^{-6}$ to $7 \times 10^{-7}$. To address overfitting, we apply a weight decay of 0.1, and gradient norms are clipped at a maximum value of 1.0.

\subsection{Offline Reinforcement Learning}

Compared to Online Reinforcement Learning (RL), Offline RL enables the pre-preparation of training signals, which is particularly advantageous for tasks where standard answers exist but are challenging to evaluate using reward models. 
In this study, we focus on objective query domains such as mathematics, coding, instruction following, and logical reasoning, where obtaining accurate evaluations can be complex.
In the previous phase, we extensively employ strategies like execution feedback and answer matching to ensure the quality of responses.
For the current phase, we reuse that pipeline, employing the SFT model to resample responses for a new set of queries. Responses that pass our quality checks are used as positive examples, while those that fail are treated as negative examples for Direct Preference Optimization (DPO) training~\citep{rafailov2024direct}.
To further enhance the reliability and accuracy of the training signals, we make use of both human and automated review processes~\citep{cao2024towards}. This dual approach ensures that the training data is not only learnable but also aligned with human expectations.
Ultimately, we construct a dataset consisting of approximately 150,000 training pairs. 
The model is then trained for one epoch using the Online Merging Optimizer \citep{lu2024online}, with a learning rate of $7 \times 10^{-7}$.

\subsection{Online Reinforcement Learning}

To develop a robust reward model for online RL, we adhere to a set of carefully defined labeling criteria. Those criteria ensure that the responses generated by the model are not only high-quality but also aligned with ethical and user-centric standards~\citep{wang2024secrets}. The specific guidelines for data labeling are as follows:

\begin{itemize}
    \item \textbf{Truthfulness:} Responses must be grounded in factual accuracy, faithfully reflecting the provided context and instructions. The model should avoid generating information that is false or unsupported by the given data.
    \item \textbf{Helpfulness:} The model's output should be genuinely useful, addressing the user's query effectively while providing content that is positive, engaging, educational, and relevant. It should follow the given instructions precisely and offer value to the user.
    \item \textbf{Conciseness:} Responses should be succinct and to the point, avoiding unnecessary verbosity. The goal is to convey information clearly and efficiently without overwhelming the user with excessive detail.
    \item \textbf{Relevance:} All parts of the response should be directly related to the user's query, dialogue history, and the assistant's context. The model should tailor its output to ensure it is perfectly aligned with the user's needs and expectations.
    \item \textbf{Harmlessness:}  The model must prioritize user safety by avoiding any content that could lead to illegal, immoral, or harmful behavior. It should promote ethical conduct and responsible communication at all times.
    \item \textbf{Debiasing:} The model should produce responses that are free from bias, including but not limited to gender, race, nationality, and politics. It should treat all topics equally and fairly, adhering to widely accepted moral and ethical standards.
\end{itemize}

The queries utilized to train the reward model are drawn from two distinct datasets: publicly available open-source data and a proprietary query set characterized by higher complexity. Responses are generated from checkpoints of the Qwen models, which have been fine-tuned using different methods—SFT, DPO, and RL—at various stages of training. To introduce diversity, those responses are sampled at different temperature settings. Preference pairs are created through both human and automated labeling processes, and the training data for DPO is also integrated into this dataset.

In our online reinforcement learning (RL) framework, we employ Group Relative Policy Optimization (GRPO,~\citealp{deepseekmath}). The query set utilized for training the reward model is identical to the one used in the RL training phase. The sequence in which queries are processed during training is determined by the variance of their response scores, as evaluated by the reward model. Specifically, queries with higher variance in response scores are prioritized to ensure more effective learning.
We sample 8 responses for each query. All models are trained with a 2048 global batch size and 2048 samples in each episode, considering a pair of queries and responses as a sample.

\subsection{Long Context Fine-tuning}

To further extend the context length of Qwen2.5-Turbo, we introduce longer SFT examples during post-training, enabling it to better align with human preference in long queries.

In the SFT phase, we employ a two-stage approach. In the first stage, the model is fine-tuned exclusively using short instructions, each containing up to 32,768 tokens. This stage uses the same data and training steps as those employed for the other Qwen2.5 models, ensuring strong performance on short tasks. In the second stage, the fine-tuning process combines both short instructions (up to 32,768 tokens) and long instructions (up to 262,144 tokens). This hybrid approach effectively enhances the model's instruction-following ability in long context tasks while maintaining its performance on short tasks.

During the RL stage, we use a training strategy similar to that used for the other Qwen2.5 models, focusing solely on short instructions. This design choice is driven by two primary considerations: first, RL training is computationally expensive for long context tasks; second, there is currently a scarcity of reward models that provide suitable reward signals for long context tasks. Additionally, we find that adopting RL on short instructions alone can still significantly enhance the model's alignment with human preferences in long context tasks.

\section{Evaluation}
\label{sec:experiment}

The base models produced by pre-training and the instruction-tuned models produced by post-training are evaluated accordingly with a comprehensive evaluation suite, including both commonly-used open benchmarks and skill-oriented in-house datasets.
The evaluation suite is designed to be primarily automatic with minimal human interaction.

To prevent test data leakage, we exclude potentially contaminated data using n-gram matching when constructing the pre-training and post-training datasets. Following the criteria used in Qwen2, a training sequence $\mathbf{s}_t$ is removed from the training data if there exists a test sequence $\mathbf{s}_e$ such that the length of the longest common subsequence (LCS) between tokenized $\mathbf{s}_t$ and $\mathbf{s}_e$ satisfies both $|\text{LCS}(\mathbf{s}_t, \mathbf{s}_e)| \geq 13$ and $|\text{LCS}(\mathbf{s}_t, \mathbf{s}_e)| \geq 0.6 \times \min(|\mathbf{s}_t|, |\mathbf{s}_e|)$.

\subsection{Base Models}

We conduct comprehensive evaluations of the base language models of the Qwen2.5 series. 
The evaluation of base models primarily emphasizes their performance in natural language understanding, general question answering, coding, mathematics, scientific knowledge, reasoning, and multilingual capabilities.

The evaluation datasets include:

\paragraph{General Tasks} MMLU~\citep{mmlu} (5-shot), MMLU-Pro~\citep{mmlupro} (5-shot), MMLU-redux~\citep{mmluredux} (5-shot), BBH~\citep{bbh} (3-shot), ARC-C~\citep{arc} (25-shot), TruthfulQA~\citep{truthfulqa} (0-shot), Winogrande~\citep{winogrande} (5-shot), HellaSwag~\citep{hellaswag} (10-shot).

\paragraph{Mathematics \& Science Tasks} GPQA~\citep{gpqa} (5-shot), Theorem QA~\citep{theoremqa} (5-shot), GSM8K~\citep{gsm8k} (4-shot), MATH~\citep{math} (4-shot).

\paragraph{Coding Tasks} HumanEval~\citep{humaneval} (0-shot), HumanEval+~\citep{evalplus}(0-shot), MBPP~\citep{mbpp} (0-shot), MBPP+~\citep{evalplus} (0-shot), MultiPL-E~\citep{multiple} (0-shot) (Python, C++, JAVA, PHP, TypeScript, C\#, Bash, JavaScript).

\paragraph{Multilingual Tasks} We group them into four categories: (a) Exam: M3Exam (5-shot, we only choose examples that require no image), IndoMMLU~\citep{koto-etal-2023-indommlu} (3-shot), ruMMLU~\citep{rummlu-mera} (5-shot), and translated MMLU~\citep{msift} (5-shot on Arabic, Spanish, French, Portuguese, German, Italian, Japanese, and Korean); (b) Understanding: BELEBELE~\citep{belebele} (5-shot), XCOPA~\citep{xcopa} (5-shot), XWinograd~\citep{xwinograd} (5-shot), XStoryCloze~\citep{xstory_cloze} (0-shot) and PAWS-X~\citep{paws-x} (5-shot); (c) Mathematics: MGSM~\citep{flores} (8-shot CoT); and (d) Translation: Flores-101~\citep{flores} (5-shot).

\begin{table}[tbp]
\centering
\caption{\textbf{Performance of the 70B+ base models and Qwen2.5-Plus.}}
\label{tab:base-70B}
\small %
\setlength{\tabcolsep}{3pt} %
\renewcommand{\arraystretch}{0.9} 

\begin{tabular}{@{}lcccccc@{}}
\toprule
\textbf{Datasets} & \textbf{Llama-3-70B} & \textbf{Mixtral-8x22B} & \textbf{Llama-3-405B} & \textbf{Qwen2-72B} & \textbf{Qwen2.5-72B} & \textbf{Qwen2.5-Plus} \\
\midrule
\multicolumn{7}{c}{\textit{General Tasks}} \\
\midrule
MMLU & 79.5 & 77.8 & 85.2 & 84.2 & \textbf{86.1} & 85.4 \\
MMLU-Pro & 52.8 & 51.6 & 61.6 & 55.7 & 58.1 & \textbf{64.0} \\
MMLU-redux & 75.0 & 72.9 & - & 80.5 & \textbf{83.9} & 82.8 \\
BBH & 81.0 & 78.9 & 85.9 & 82.4 & \textbf{86.3} & 85.8 \\
ARC-C & 68.8 & 70.7 & - & 68.9 & \textbf{72.4} & 70.9 \\
TruthfulQA & 45.6 & 51.0 & - & 54.8 & \textbf{60.4} & 55.3 \\
WindoGrande & 85.3 & 85.0 & \textbf{86.7} & 85.1 & 83.9 & 85.5 \\
HellaSwag & 88.0 & 88.7 & - & 87.3 & 87.6 & \textbf{89.2} \\
\midrule
\multicolumn{7}{c}{\textit{Mathematics \& Science Tasks}} \\
\midrule
GPQA & 36.3 & 34.3 & - & 37.4 & \textbf{45.9} & 43.9\\
TheoremQA & 32.3 & 35.9 & - & 42.8 & 42.4 & \textbf{48.5} \\
MATH & 42.5 & 41.7 & 53.8 & 50.9 & 62.1 & \textbf{64.4} \\
MMLU-stem & 73.7 & 71.7 & - & 79.6 & \textbf{82.7} & 81.2 \\
GSM8K & 77.6 & 83.7 & 89.0 & 89.0 & 91.5 & \textbf{93.0} \\
\midrule
\multicolumn{7}{c}{\textit{Coding Tasks}} \\
\midrule
HumanEval & 48.2 & 46.3 & \textbf{61.0} & 64.6 & 59.1 & 59.1 \\
HumanEval+ & 42.1 & 40.2 & - & \textbf{56.1} & 51.2 & 52.4\\
MBPP & 70.4 & 71.7 & 73.0 & 76.9 & \textbf{84.7} & 79.7\\
MBPP+ & 58.4 & 58.1 & - & 63.9 & \textbf{69.2} & 66.9 \\
MultiPL-E & 46.3 & 46.7 & - & 59.6 & 60.5 & \textbf{61.0} \\
\midrule
\multicolumn{7}{c}{\textit{Multilingual Tasks}} \\
\midrule
Multi-Exam & 70.0 & 63.5 & - & 76.6 & \textbf{78.7} & 78.5 \\
Multi-Understanding & 79.9 & 77.7 & - & 80.7 & \textbf{89.6} & 89.2 \\
Multi-Mathematics & 67.1 & 62.9 & - & 76.0 & 76.7 & \textbf{82.4} \\
Multi-Translation & 38.0 & 23.3 & - & 37.8 & 39.0 & \textbf{40.4} \\
\bottomrule
\end{tabular}
\end{table}

For base models, we compare Qwen2.5 models with Qwen2 models and other leading open-weight models in terms of scales of parameters.

\paragraph{Qwen2.5-72B \& Qwen2.5-Plus}
We compare the base models of Qwen2.5-72B and Qwen2.5-Plus to other leading open-weight base models: Llama3-70B~\citep{llama3}, Llama3-405B~\citep{llama3}, Mixtrail-8x22B~\citep{mixtral}, and our previous 72B version, the Qwen2-72B~\citep{qwen2}. The Qwen2.5-72B base model significantly outperforms its peers in the same category across a wide range of tasks. It achieves results comparable to Llama-3-405B while utilizing only one-fifth of the parameters. Furthermore, when compared to its predecessor, Qwen2-72B, the Qwen2.5-72B shows marked improvements in nearly all benchmark evaluations, particularly excelling in general tasks, mathematics, and coding challenges. With significantly lower training and inference costs, Qwen2.5-Plus achieves very competitive performance results compared to Qwen2.5-72B and Llama3-405B, outperforming other baseline models on the Hellaswag, TheoremQA, MATH, GSM8K, MultiPL-E, Multi-Mathematics, and Multi-Translation. Moreover, Qwen2.5-Plus achieves 64.0 on MMLU-Pro, which is 5.9 points higher than Qwen2.5-72B.

\begin{table}[tbp]
\centering
\caption{\textbf{Performance of the 14B-30B+ base models and Qwen2.5-Turbo.}}
\label{tab:base-14B}
\small %
\setlength{\tabcolsep}{3pt} %
\renewcommand{\arraystretch}{0.9} 
\begin{tabular}{@{}lcccccc@{}}
\toprule
\textbf{Datasets} & \textbf{Qwen1.5-32B} & \textbf{Gemma2-27B} & \textbf{Yi-1.5-34B} & \textbf{Qwen2.5-Turbo} & \textbf{Qwen2.5-14B} & \textbf{Qwen2.5-32B} \\
\midrule
\multicolumn{7}{c}{\textit{General Tasks}} \\
\midrule
MMLU          & 74.3 & 75.2 & 77.2 & 79.5 & 79.7 & \textbf{83.3} \\
MMLU-pro      & 44.1 & 49.1 & 48.3 & \textbf{55.6} & 51.2 & 55.1 \\
MMLU-redux    & 69.0 & -    & 74.1 & 77.1 & 76.6 & \textbf{82.0} \\
BBH           & 66.8 & 74.9 & 76.4 & 76.1 & 78.2 & \textbf{84.5} \\
ARC-C         & 63.6 & \textbf{71.4} & 65.6 & 67.8 & 67.3 & 70.4 \\
TruthfulQA    & 57.4 & 40.1 & 53.9 & 56.3 & \textbf{58.4} & 57.8 \\
Winogrande    & 81.5 & 59.7 & \textbf{84.9} & 81.1 & 81.0    & 82.0 \\
Hellaswag     & 85.0 & \textbf{86.4} & 85.9 & 85.0 & 84.3    & 85.2 \\
\midrule
\multicolumn{7}{c}{\textit{Mathematics \& Science Tasks}} \\
\midrule
GPQA          & 30.8 & 34.9 & 37.4 & 41.4 & 32.8 & \textbf{48.0} \\
Theoremqa     & 28.8 & 35.8 & 40.0 & 42.1 & 43.0 & \textbf{44.1} \\
MATH          & 36.1 & 42.7 & 41.7 & 55.6 & 55.6 & \textbf{57.7} \\
MMLU-stem     & 66.5 & 71.0 & 72.6 & 77.0 & 76.4 & \textbf{80.9} \\
GSM8K         & 78.5 & 81.1 & 81.7 & 88.3 & 90.2 & \textbf{92.9} \\
\midrule
\multicolumn{7}{c}{\textit{Coding Tasks}} \\
\midrule
HumanEval     & 43.3 & 54.9 & 46.3 & 57.3 & 56.7 & \textbf{58.5} \\
HumanEval+    & 40.2 & 46.3 & 40.2 & 51.2 & 51.2 & \textbf{52.4} \\
MBPP          & 64.2 & 75.7 & 65.5 & 76.2 & 76.7 & \textbf{84.5} \\
MBPP+         & 53.9 & 60.2 & 55.4 & 63.0 & 63.2 & \textbf{67.2} \\
MultiPL-E     & 38.5 & 48.0 & 39.5 & 53.9 & 53.5 & \textbf{59.4} \\
\midrule
\multicolumn{7}{c}{\textit{Multilingual Tasks}} \\
\midrule
Multi-Exam         & 61.6 & 65.8 & 58.3 & 70.3 & 70.6 & \textbf{75.4} \\
Multi-Understanding& 76.5 & 82.2 & 73.9 & 85.3 & 85.9 & \textbf{88.4} \\
Multi-Mathematics  & 56.1 & 61.6 & 49.3 & 71.3 & 68.5 & \textbf{73.7} \\
Multi-Translation  & 33.5 & \textbf{38.7} & 30.0 & 36.8 & 36.2 & 37.3 \\
\bottomrule
\end{tabular}
\end{table}

\begin{table}[tbp]
\centering
\caption{\textbf{Performance of the 7B+ base models.}}
\label{tab:base-7B}
\small
\begin{tabular}{@{}lccccc@{}}
\toprule
\textbf{Datasets} & \textbf{Mistral-7B} & \textbf{Llama3-8B} & \textbf{Gemma2-9B} & \textbf{Qwen2-7B} & \textbf{Qwen2.5-7B} \\
\midrule
\multicolumn{6}{c}{\textit{General Tasks}} \\
\midrule
MMLU & 64.2 & 66.6 & 71.3 & 70.3 & \textbf{74.2} \\
MMLU-pro & 30.9 & 35.4 & 44.7 & 40.1 & \textbf{45.0} \\
MMLU-redux & 58.1 & 61.6 & 67.9 & 68.1 & \textbf{71.1} \\
BBH & 56.1 & 57.7 & 68.2 & 62.3 & \textbf{70.4} \\
ARC-C & 60.0 & 59.3 & \textbf{68.2} & 60.6 & 63.7 \\
TruthfulQA & 42.2 & 44.0 & 45.3 & 54.2 & \textbf{56.4} \\
Winogrande & 78.4 & 77.4 & \textbf{79.5} & 77.0 & 75.9 \\
HellaSwag & \textbf{83.3} & 82.1 & 81.9 & 80.7 & 80.2 \\
\midrule
\multicolumn{6}{c}{\textit{Mathematics \& Science Tasks}} \\
\midrule
GPQA & 24.7 & 25.8 & 32.8 & 30.8 & \textbf{36.4} \\
TheoremQA & 19.2 & 22.1 & 28.9 & 29.6 & \textbf{36.0} \\
MATH & 10.2 & 20.5 & 37.7 & 43.5 & \textbf{49.8} \\
MMLU-stem & 50.1 & 55.3 & 65.1 & 64.2 & \textbf{72.3} \\
GSM8K & 36.2 & 55.3 & 70.7 & 80.2 & \textbf{85.4} \\
\midrule
\multicolumn{6}{c}{\textit{Coding Tasks}} \\
\midrule
HumanEval & 29.3 & 33.5 & 37.8 & 51.2 & \textbf{57.9} \\
HumanEval+ & 24.4 & 29.3 & 30.5 & 43.3 & \textbf{50.6} \\
MBPP & 51.1 & 53.9 & 62.2 & 64.2 & \textbf{74.9} \\
MBPP+ & 40.9 & 44.4 & 50.6 & 51.9 & \textbf{62.9} \\
MultiPL-E & 29.4 & 22.6 & 34.9 & 41.0 & \textbf{50.3} \\
\midrule
\multicolumn{6}{c}{\textit{Multilingual Tasks}} \\
\midrule
Multi-Exam & 47.1 & 52.3 & \textbf{61.2} & 59.2 & 59.4 \\
Multi-Understanding & 63.3 & 68.6 & 78.3 & 72.0 & \textbf{79.3} \\
Multi-Mathematics & 26.3 & 36.3 & 53.0 & 57.5 & \textbf{57.8} \\
Multi-Translation & 23.3 & 31.9 & \textbf{36.5} & 31.5 & 32.4 \\
\bottomrule
\end{tabular}
\end{table}

\paragraph{Qwen2.5-14B/32B \& Qwen2.5-Turbo}
The evaluation of the Qwen2.5-Turbo, Qwen2.5-14B, and 32B models is compared against baselines of similar sizes. 
These baselines include Yi-1.5-34B~\citep{yi}, Gemma2-27B~\citep{gemma2}, and Qwen1.5-32B~\citep{qwen1.5}.
The results are shown in Table~\ref{tab:base-14B}.
The Qwen2.5-14B model demonstrates a solid performance across various tasks, particularly excelling in general tasks like MMLU and BBH, where it achieves scores of 79.7 and 78.2, outcompeting competitors of larger sizes. Meanwhile, Qwen2.5-32B, in particular, showcases exceptional capabilities, often surpassing larger models of similar model sizes. Notably, it outperforms its predecessor Qwen1.5-32B significantly, especially in challenging areas such as mathematics and coding, with notable scores of 57.7 in MATH and 84.5 in MBPP. For Qwen2.5-Turbo, although its training cost and inference cost are significantly smaller than those of Qwen2.5-14B, it achieves comparable results, where its MMLU-Pro score is even better than that of Qwen2.5-32B.

\paragraph{Qwen2.5-7B}
For 7B-level models, we focus on comparing Qwen2.5-7B with other leading 7B+ models, including Mistral-7B~\citep{mistral}, Llama3-8B~\citep{llama3}, Gemma2-9B~\citep{gemma2}, and our predecessor, Qwen2-7B~\citep{qwen2}.
The results can be found in Table~\ref{tab:base-7B}. Note that the non-embedding parameters of Qwen2-7B and Qwen2.5-7B are only 6.5B, while that of Gemma2-9B is 8.2B.
The Qwen2.5-7B model surpasses its predecessors and counterparts in numerous benchmarks, despite having fewer non-embedding parameters. It demonstrates significant improvements across various tasks, achieving 74.2 on general benchmarks like MMLU~\citep{mmlu}, 49.8 on math challenges such as MATH~\citep{math}, and 57.9 on coding tasks like HumanEval~\citep{humaneval}.

\begin{table}[t]
\centering
\caption{\textbf{Performance of the smaller base models.}}
\label{tab:base-small}
\small %
\setlength{\tabcolsep}{3pt} %
\renewcommand{\arraystretch}{0.9} %
\begin{tabular}{@{}lcccccc@{}}
\toprule
\textbf{Datasets} & \textbf{Qwen2-0.5B} & \textbf{Qwen2.5-0.5B} & \textbf{Qwen2-1.5B} & \textbf{Qwen2.5-1.5B} & \textbf{Gemma2-2.6B} & \textbf{Qwen2.5-3B} \\
\midrule
\multicolumn{7}{c}{\textit{General Tasks}} \\
\midrule
MMLU & 44.3 & 47.5 & 55.9 & 60.9 & 52.2 & \textbf{65.6} \\
MMLU-pro & 14.7 & 15.7 & 21.6 & 28.5 & 23.0 & \textbf{34.6} \\
MMLU-redux & 40.7 & 45.1 & 51.8 & 58.5 & 50.9 & \textbf{63.7} \\
BBH & 18.2 & 20.3 & 36.5 & 45.1 & 41.9 & \textbf{56.3} \\
ARC-C & 31.0 & 35.6 & 43.7 & 54.7 & 55.7 & \textbf{56.5} \\
TruthfulQA & 39.7 & 40.2 & 45.9 & 46.6 & 36.2 & \textbf{48.9} \\
Winogrande & 56.9 & 56.3 & 65.0 & 65.0 & \textbf{71.5} & 71.1 \\
Hellaswag & 49.1 & 52.1 & 67.0 & 67.9 & \textbf{74.6} & \textbf{74.6} \\
\midrule
\multicolumn{7}{c}{\textit{Mathematics \& Science Tasks}} \\
\midrule
GPQA & \textbf{29.8} & 24.8 & 20.7 & 24.2 & 25.3 & 26.3 \\
TheoremQA & 9.6 & 16.0 & 14.8 & 22.1 & 15.9 & \textbf{27.4} \\
MATH & 11.2 & 19.5 & 21.6 & 35.0 & 18.3 & \textbf{42.6} \\
MMLU-STEM & 27.5 & 39.8 & 42.7 & 54.8 & 45.8 & \textbf{62.5} \\
GSM8K & 36.4 & 41.6 & 46.9 & 68.5 & 30.3 & \textbf{79.1} \\
\midrule
\multicolumn{7}{c}{\textit{Coding Tasks}} \\
\midrule
HumanEval & 22.6 & 30.5 & 34.8 & 37.2 & 19.5 & \textbf{42.1} \\
HumanEval+ & 18.9 & 26.8 & 29.9 & 32.9 & 15.9 & \textbf{36.0} \\
MBPP & 33.1 & 39.3 & 46.9 & \textbf{60.2} & 42.1 & 57.1 \\
MBPP+ & 27.6 & 33.8 & 37.6 & \textbf{49.6} & 33.6 & 49.4 \\
MultiPL-E & 16.3 & 18.9 & 27.9 & 33.1 & 17.6 & \textbf{41.2} \\
\midrule
\multicolumn{7}{c}{\textit{Multilingual Tasks}} \\
\midrule
Multi-Exam & 29.4 & 30.8 & 43.1 & 47.9 & 38.1 & \textbf{54.6} \\
Multi-Understanding & 40.4 & 41.0 & 50.7 & 65.1 & 46.8 & \textbf{76.6} \\
Multi-Mathematics & 7.8 & 13.5 & 21.3 & 37.5 & 18.2 & \textbf{48.9} \\
Multi-Translation & 14.1 & 15.3 & 23.8 & 25.0 & 26.9 & \textbf{29.3} \\
\bottomrule
\end{tabular}
\end{table}

\paragraph{Qwen2.5-0.5B/1.5B/3B}
For edge-side models, we compare Qwen2.5-0.5B, 1.5B, and 3B against established baselines: Qwen2-0.5B/1.5B~\citep{qwen2} and Gemma2-2.6B~\citep{gemma2}. The results are given in Table~\ref{tab:base-small}. Qwen2.5-0.5B, 1.5B, and 3B continue to maintain strong performance across nearly all benchmarks. Notably, the Qwen2.5-0.5B model outperforms the Gemma2-2.6B on various math and coding tasks.

\subsection{Instruction-tuned Model}

To critically evaluate instruction-tuned models, we adopt a multifaceted approach. 
Foundational skills and human preferences are assessed using open datasets and benchmarks. 
Additionally, our detailed in-house evaluations delve deeper into the models'competencies in key areas and multilingualism. 
A particular focus is placed on assessing long-context capability.
The subsequent sections outline the evaluation methods and present the results.

\subsubsection{Open Benchmark Evaluation}

To comprehensively evaluate the quality of instruction-tuned models, we compile automatic and human evaluation to assess the capabilities and human preference. 
For the evaluation of basic capabilities, we apply similar datasets in the pre-trained model evaluation, which target on natural language understanding, coding, mathematics, and reasoning. 
Specifically, we evaluate on MMLU-Pro, MMLU-redux and LiveBench 0831~\citep{livebench} for general evaluation, GPQA, GSM8K and MATH for science and mathematics, HumanEval, MBPP, MultiPL-E and LiveCodeBench 2305-2409~\citep{livecodebench} for coding, 
IFEval~\citep{ifeval}\footnote{For simplicity, we report the results of the subset \textit{strict-prompt}.} for instruction following.
Additionally, we assess the performance of human preference alignment and instruction following by evaluating on benchmarks including MT-Bench~\citep{mtbench} and Arena-Hard~\citep{arena-hard}.

\begin{table}[tbp]

\centering

\caption{\textbf{Performance of the 70B+ Instruct models and Qwen2.5-Plus.}}

\label{tab:70b_instruct}
\small
\setlength{\tabcolsep}{2.6pt}

\begin{tabular}{@{}lccccc@{}}

\toprule

\textbf{Datasets}  & \textbf{Llama-3.1-70B} & \textbf{Llama-3.1-405B} & \textbf{Qwen2-72B} & \textbf{Qwen2.5-72B} & \textbf{Qwen2.5-Plus} \\

\midrule

\multicolumn{6}{c}{\textit{General Tasks}} \\
\midrule

MMLU-Pro  & 66.4 & \textbf{73.3} & 64.4 & 71.1 & 72.5 \\

MMLU-redux  & 83.0 & 86.2 & 81.6 & \textbf{86.8} & 86.3\\

LiveBench 0831  & 46.6 & 53.2 & 41.5 & 52.3 & \textbf{54.6}\\

\midrule
\multicolumn{6}{c}{\textit{Mathematics \& Science Tasks}} \\
\midrule

GPQA   & 46.7 & \textbf{51.1} & 42.4 & 49.0 & 49.7\\

MATH   & 68.0 & 73.8 & 69.0 & 83.1 & \textbf{84.7}\\

GSM8K  & 95.1 & \textbf{96.8} & 93.2 & 95.8 & 96.0\\

\midrule
\multicolumn{6}{c}{\textit{Coding Tasks}} \\
\midrule

HumanEval  & 80.5 & \textbf{89.0} & 86.0 & 86.6 & 87.8\\

MBPP  & 84.2 & 84.5 & 80.2 & \textbf{88.2} & 85.5\\

MultiPL-E  & 68.2 & 73.5 & 69.2 & 75.1 & \textbf{77.0}  \\

LiveCodeBench  & 32.1 & 41.6 & 32.2 & \textbf{55.5} & 51.4\\

\midrule
\multicolumn{6}{c}{\textit{Alignment Tasks}} \\
\midrule

IFEval  & 83.6 & 86.0 & 77.6 & 84.1 & \textbf{86.3}\\

Arena-Hard  & 55.7 & 69.3 & 48.1 & 81.2 & \textbf{81.4}\\

MTbench  & 8.79 & 9.08 & 9.12 & \textbf{9.35} & 9.30\\

\bottomrule

\end{tabular}

\end{table}

\begin{table}[tbp]

\centering

\caption{\textbf{Performance of the 14B-30B+ instruction-tuned models and Qwen2.5-Turbo.}}

\label{tab:14b-instruct}
\small
\setlength{\tabcolsep}{2.6pt}

\begin{tabular}{@{}lcccccc@{}}

\toprule

\textbf{Datasets} & \textbf{Qwen2-57BA14B} & \textbf{Gemma2-27B} & \textbf{GPT4o-mini} & \textbf{Qwen2.5-Turbo} & \textbf{Qwen2.5-14B} & \textbf{Qwen2.5-32B} \\

\midrule

\multicolumn{7}{c}{\textit{General Tasks}} \\

\midrule

MMLU-Pro & 52.8 & 55.5 & 63.1 & 64.5 & 63.7 & \textbf{69.0} \\

MMLU-redux & 72.6 & 75.7 & 81.5 & 81.7 & 80.0 & \textbf{83.9} \\

LiveBench 0831 & 31.1 & 39.6 & 43.3 & 42.3 & 44.4 & \textbf{50.7} \\

\midrule

\multicolumn{7}{c}{\textit{Mathematics \& Science Tasks}} \\

\midrule

GPQA & 34.3 & 38.4 & 40.2 & 42.3 & 45.5 & \textbf{49.5} \\

MATH & 49.1 & 54.4 & 70.2 & 81.1 & 80.0 & \textbf{83.1} \\

GSM8K & 85.3 & 90.4 & 93.2 & 93.8 & 94.8 & \textbf{95.9} \\

\midrule

\multicolumn{7}{c}{\textit{Coding Tasks}} \\

\midrule

HumanEval & 79.9 & 78.7 & \textbf{88.4} & 86.6 & 83.5 & \textbf{88.4} \\

MBPP & 70.9 & 81.0 & \textbf{85.7} & 82.8 & 82.0 & 84.0 \\

MultiPL-E & 66.4 & 67.4 & 75.0 & 73.7 & 72.8 & \textbf{75.4} \\

LiveCodeBench & 22.5 & - & 40.7 & 37.8 & 42.6 & \textbf{51.2} \\

\midrule

\multicolumn{7}{c}{\textit{Alignment Tasks}} \\

\midrule

IFEval & 59.9 & 77.1 & 80.4 & 76.3 & \textbf{81.0} & 79.5 \\

Arena-Hard & 17.8 & 57.5 & \textbf{74.9} & 67.1 & 68.3 & 74.5 \\

MTbench & 8.55 & 9.10 & - & 8.81 & 8.88 & \textbf{9.20} \\

\bottomrule

\end{tabular}

\end{table}

\paragraph{Qwen2.5-72B-Instruct \& Qwen2.5-Plus} 
As shown in Table~\ref{tab:70b_instruct}, we compare Qwen2.5-72B-Instruct and Qwen2.5-Plus to other leading open-weight instrution-tuned models: Llama3.1-70B-Instruct~\citep{llama3}, Llama3.1-405B-Instruct~\citep{llama3}, and our previous 72B version, Qwen2-72B-Instruct~\citep{qwen2}. 
The Qwen2.5-72B-Instruct model delivers exceptional performance, even surpassing the larger Llama-3.1-405B-Instruct in several critical benchmarks including MMLU-redux, MATH, MBPP, MultiPL-E, LiveCodeBench, Arena-Hard and MTBench.
Moreover, Qwen2.5-Plus outperforms Qwen2.5-72B-Instruct on 9 out of 13 benchmarks.

\paragraph{Qwen2.5-14B/32B-Instruct \& Qwen2.5-Turbo}
The performance of the Qwen2.5-Turbo, Qwen2.5-14B-Instruct, and Qwen2.5-32B-Instruct models is evaluated and compared against baselines of similar sizes. The baselines include GPT4o-mini, Gemma2-27B-IT~\citep{gemma2}, and Qwen2-57BA14B-Instruct~\citep{qwen2}. The results are summarized in Table~\ref{tab:14b-instruct}.
The Qwen2.5-32B-Instruct model exhibits superior performance across most tasks when compared to other models of similar size. Notably, our open-weight Qwen2.5-14B-Instruct model delivers competitive results across all benchmarks, rivaling those of GPT-4o-mini.
Despite its significantly lower training and inference costs, the Qwen2.5-Turbo model outperforms Qwen2.5-14B-Instruct on eight out of ten benchmarks. This demonstrates that Qwen2.5-Turbo achieves remarkable efficiency and effectiveness, making it a compelling choice for resource-constrained environments.

\begin{table}[t]

\centering

\caption{\textbf{Performance of the 7B+ instruction-tuned models.}}

\label{tab:7b-instruct}
\small
\setlength{\tabcolsep}{2.6pt}

\begin{tabular}{@{}lcccc@{}}

\toprule

\textbf{Datasets} & \textbf{Gemma2-9B} & \textbf{Llama3.1-8B} & \textbf{Qwen2-7B} & \textbf{Qwen2.5-7B} \\

\midrule

\multicolumn{5}{c}{\textit{General Tasks}} \\

\midrule

MMLU-Pro & 52.1 & 48.3 & 44.1 & \textbf{56.3} \\

MMLU-redux & 72.8 & 67.2 & 67.3 & \textbf{75.4} \\

LiveBench 0831 & 30.6 & 26.7 & 29.2 & \textbf{35.9} \\

\midrule

\multicolumn{5}{c}{\textit{Mathematics \& Science Tasks}} \\

\midrule

GPQA & 32.8 & 32.8 & 34.3 & \textbf{36.4} \\

MATH & 44.3 & 51.9 & 52.9 & \textbf{75.5} \\

GSM8K & 76.7 & 84.5 & 85.7 & \textbf{91.6} \\

\midrule

\multicolumn{5}{c}{\textit{Coding Tasks}} \\

\midrule

HumanEval & 68.9 & 72.6 & 79.9 & \textbf{84.8} \\

MBPP & 74.9 & 69.6 & 67.2 & \textbf{79.2} \\

MultiPL-E & 53.4 & 50.7 & 59.1 & \textbf{70.4} \\

LiveCodeBench & 18.9 & 8.3 & 23.9 & \textbf{28.7} \\

\midrule

\multicolumn{5}{c}{\textit{Alignment Tasks}} \\

\midrule

IFEval & 70.1 & \textbf{75.9} & 54.7 & 71.2 \\

Arena-Hard & 41.6 & 27.8 & 25.0 & \textbf{52.0} \\

MTbench & 8.49 & 8.23 & 8.26 & \textbf{8.75} \\

\bottomrule

\end{tabular}

\end{table}

\begin{table}[t]

\centering

\caption{\textbf{Performance comparison of 2B-4B instruction-tuned models.}}

\label{tab:4b-instruct}
\small
\setlength{\tabcolsep}{2.6pt}

\begin{tabular}{@{}lcccc@{}}

\toprule

\textbf{Datasets} & \textbf{Gemma2-2B} & \textbf{Phi3.5-Mini} & \textbf{MiniCPM3-4B} & \textbf{Qwen2.5-3B} \\

\midrule

Non-Emb Params & 2.0B & 3.6B & 4.0B & 2.8B \\

\midrule

\multicolumn{5}{c}{\textit{General Tasks}} \\

\midrule

MMLU-Pro & 26.7 & \textbf{47.5} & 43.0 & 43.7 \\

MMLU-redux & 51.9 & \textbf{67.7} & 59.9 & 64.4 \\

LiveBench 0831 & 20.1 & 27.4 & \textbf{27.6} & 26.8 \\

\midrule

\multicolumn{5}{c}{\textit{Mathematics \& Science Tasks}} \\

\midrule

GPQA & 29.3 & 27.2 & \textbf{31.3} & 30.3 \\

MATH & 26.6 & 48.5 & 46.6 & \textbf{65.9} \\

GSM8K & 63.2 & 86.2 & 81.1 & \textbf{86.7} \\

\midrule

\multicolumn{5}{c}{\textit{Coding Tasks}} \\

\midrule

HumanEval & 68.9 & 72.6 & \textbf{74.4} & \textbf{74.4} \\

MBPP & \textbf{74.9} & 63.2 & 72.5 & 72.7 \\

MultiPL-E & 30.5 & 47.2 & 49.1 & \textbf{60.2} \\

LiveCodeBench & 5.8 & 15.8 & \textbf{23.8} & 19.9 \\

\midrule

\multicolumn{5}{c}{\textit{Alignment Tasks}} \\

\midrule

IFEval & 51.0 & 52.1 & \textbf{68.4} & 58.2 \\

\bottomrule

\end{tabular}

\end{table}

\paragraph{Other Instruction-tuned Models} 
As illustrated in Table~\ref{tab:7b-instruct}, the Qwen2.5-7B-Instruct model significantly outperforms its competitors, Gemma2-9B-IT and Llama3.1-8B-Instruct, across all tasks except IFEval. Notably, Qwen2.5-7B-Instruct exhibits clear advantages in mathematics (MATH: 75.5) and coding (HumanEval: 84.8).
For the edge-side instruction models, the Qwen2.5-3B-Instruct model, despite having fewer parameters than both the Phi3.5-mini-instruct~\citep{phi3} and MiniCPM3-4B-Instruct~\citep{minicpm} models, surpasses them in mathematics and coding tasks, as shown in Table~\ref{tab:4b-instruct}. Additionally, it delivers competitive results in language understanding.
The Qwen2.5-1.5B-Instruct and Qwen2.5-0.5B-Instruct models have also seen substantial performance improvements over their previous versions, as detailed in Table~\ref{tab:2b-instruct}. These enhancements make them particularly well-suited for edge-side applications in highly resource-constrained environments.

\begin{table}[t]

\centering

\caption{\textbf{Performance comparison of 0.5B-1.5B instruction-tuned models.}}

\label{tab:2b-instruct}
\small
\setlength{\tabcolsep}{2.6pt}

\begin{tabular}{@{}lcccc@{}}

\toprule

\textbf{Datasets} & \textbf{Qwen2-0.5B} & \textbf{Qwen2.5-0.5B} & \textbf{Qwen2-1.5B} & \textbf{Qwen2.5-1.5B} \\

\midrule

\multicolumn{5}{c}{\textit{General Tasks}} \\

\midrule

MMLU-Pro & 14.4 & \textbf{15.0} & 22.9 & \textbf{32.4} \\

MMLU-redux & 12.9 & \textbf{24.1} & 41.2 & \textbf{50.7} \\

LiveBench & 7.4 & \textbf{12.6} & 12.4 & \textbf{18.8} \\

\midrule

\multicolumn{5}{c}{\textit{Mathematics \& Science Tasks}} \\

\midrule

GPQA & 23.7 & \textbf{29.8} & 21.2 & \textbf{29.8} \\

MATH & 13.9 & \textbf{34.4} & 25.3 & \textbf{55.2} \\

GSM8K & 40.1 & \textbf{49.6} & 61.6 & \textbf{73.2} \\

\midrule

\multicolumn{5}{c}{\textit{Coding Tasks}} \\

\midrule

HumanEval & 31.1 & \textbf{35.4} & 42.1 & \textbf{61.6} \\

MBPP & 39.7 & \textbf{49.6} & 44.2 & \textbf{63.2} \\

MultiPL-E & 20.8 & \textbf{28.5} & 38.5 & \textbf{50.4} \\

LiveCodeBench & 1.6 & \textbf{5.1} & 4.5 & \textbf{14.8} \\

\midrule

\multicolumn{5}{c}{\textit{Alignment Tasks}} \\

\midrule

IFEval & 14.6 & \textbf{27.9} & 29.0 & \textbf{42.5} \\

\bottomrule

\end{tabular}

\end{table}

\subsubsection{In-house Automatic Evaluation}

\begin{table}[tbp]
\centering
\small
\caption{\textbf{Performance Comparison on our in-house English automatic evaluation benchmark.} }
\label{tab:eval_v11_EN}
\begin{tabular}{@{}lcccccc@{}}
\toprule
\textbf{Models} & \textbf{IF} & \textbf{Knowledge} & \textbf{Comprehension} & \textbf{Coding} & \textbf{Math} & \textbf{Reasoning} \\
\midrule
\multicolumn{7}{c}{\textit{Proprietary LLMs}} \\
\midrule
GPT-4o-2024-08-06 & 83.28 & 68.08 & 76.51 & 58.05 & 52.36 & 66.45 \\
GPT-4o-2024-11-20 & 80.06 & 65.25 & 79.07 & 60.19 & 49.74 & 67.07 \\
Claude3.5-sonnet-2024-10-22 & 84.22 & 74.61 & 79.02 & 67.17 & 48.67 & 70.20\\
\midrule
\multicolumn{7}{c}{\textit{Qwen2 Series}} \\
\midrule
Qwen2-0.5B-Instruct & 18.33 & 18.59 & 30.64 & 5.42 & 13.16 & 32.03 \\
Qwen2-1.5B-Instruct & 29.42 & 29.23 & 45.81 & 17.02 & 20.34 & 38.86 \\
Qwen2-7B-Instruct & 50.47 & 44.79 & 58.04 & 43.04 & 38.31 & 50.25 \\
Qwen2-72B-Instruct & 76.08 & 59.49 & 72.19 & 48.95 & 48.07 & 60.33 \\
\midrule
\multicolumn{7}{c}{\textit{Llama-3.1 Series}} \\
\midrule
Llama-3.1-70B-Instruct & 81.33 & 63.42 & 69.29 & 55.96 & 48.00 & 63.18 \\
Llama-3.1-405B-Instruct & 83.33 & 67.10 & 75.55 & 58.14 & 47.09 & 64.74\\
\midrule
\multicolumn{7}{c}{\textit{Qwen2.5 Series}} \\
\midrule
Qwen2.5-0.5B-Instruct & 33.35 & 30.29 & 29.78 & 15.41 & 26.29 & 36.13 \\
Qwen2.5-1.5B-Instruct & 40.25 & 41.19 & 47.69 & 26.19 & 40.99 & 42.23 \\
Qwen2.5-3B-Instruct & 60.60 & 46.11 & 57.98 & 41.43 & 49.38 & 49.80 \\
Qwen2.5-7B-Instruct & 70.01 & 52.74 & 62.69 & 48.41 & 56.93 & 54.69 \\
Qwen2.5-14B-Instruct & 74.17 & 59.78 & 69.11 & 52.68 & 59.68 & 62.51 \\
Qwen2.5-Turbo        & 72.76 & 58.56 & 68.70 & 54.48 & 57.77 & 61.06 \\
Qwen2.5-32B-Instruct & 76.79 & 64.08 & 71.28 & 58.90 & 60.97 & 65.49 \\
Qwen2.5-72B-Instruct & 82.65 & 66.09 & 74.43 & 60.41 & 59.73 & 65.90 \\
Qwen2.5-Plus         & 83.18 & 68.41 & 79.35 & 59.58 & 62.52 & 66.92 \\
\bottomrule
\end{tabular}
\end{table}

\begin{table}[tbp]
\centering
\small
\caption{\textbf{Performance Comparison on our in-house Chinese automatic evaluation benchmark.} }
\label{tab:eval_v11_ZH}
\begin{tabular}{@{}lcccccc@{}}
\toprule
\textbf{Models} & \textbf{IF} & \textbf{Knowledge} & \textbf{Comprehension} & \textbf{Coding} & \textbf{Math} & \textbf{Reasoning} \\
\midrule
\multicolumn{7}{c}{\textit{Proprietary LLMs}} \\
\midrule
GPT-4o-2024-08-06 & 42.50 & 68.55 & 80.11 & 61.53 & 61.74 & 56.88 \\
GPT-4o-2024-11-20 & 42.71 & 71.29 & 83.04 & 62.39 & 66.04 & 62.04 \\
Claude3.5-sonnet-2024-10-22 & 49.25 & 72.09 & 82.16 & 66.00 & 63.71 & 66.60\\
\midrule
\multicolumn{7}{c}{\textit{Qwen2 Series}} \\
\midrule
Qwen2-0.5B-Instruct & 4.69 & 40.43 & 39.13 & 9.85 & 14.07 & 32.73 \\
Qwen2-1.5B-Instruct & 6.81 & 51.54 & 46.89 & 14.14 & 24.57 & 35.19 \\
Qwen2-7B-Instruct & 16.83 & 65.95 & 60.30 & 37.05 & 50.52 & 44.96 \\
Qwen2-72B-Instruct & 31.98 & 74.96 & 75.49 & 41.57 & 65.55 & 58.19 \\
\midrule
\multicolumn{7}{c}{\textit{Llama-3.1 Series}} \\
\midrule
Llama-3.1-70B-Instruct & 28.96 & 57.41 & 67.24 & 54.82 & 41.18 & 52.42 \\
Llama-3.1-405B-Instruct & 30.39 & 63.79 & 72.27 & 60.73 & 46.05 & 55.88\\
\midrule
\multicolumn{7}{c}{\textit{Qwen2.5 Series}} \\
\midrule
Qwen2.5-0.5B-Instruct & 6.12 & 39.13 & 42.97 & 9.60 & 24.03 & 33.72 \\
Qwen2.5-1.5B-Instruct & 7.38 & 48.68 & 49.69 & 22.96 & 37.30 & 39.17 \\
Qwen2.5-3B-Instruct & 16.50 & 57.18 & 62.55 & 29.88 & 51.64 & 39.57 \\
Qwen2.5-7B-Instruct & 26.64 & 65.77 & 67.55 & 39.56 & 61.06 & 49.70 \\
Qwen2.5-14B-Instruct & 26.87 & 70.28 & 76.96 & 49.78 & 67.01 & 56.41 \\
Qwen2.5-Turbo        & 32.94 & 72.93 & 74.37 & 51.92 & 66.08 & 53.30 \\
Qwen2.5-32B-Instruct & 32.64 & 74.70 & 79.46 & 54.45 & 67.86 & 60.19 \\
Qwen2.5-72B-Instruct & 37.22 & 75.86 & 78.85 & 56.71 & 68.39 & 63.02 \\
Qwen2.5-Plus         & 46.15 & 72.07 & 82.64 & 58.48 & 69.96 & 62.98 \\
\bottomrule
\end{tabular}
\end{table}

Despite the availability of several open benchmark datasets for evaluation, we believe that these are insufficient to fully capture the capabilities of LLMs.
To address this, we have developed a series of in-house datasets designed to assess various aspects of model performance, including knowledge understanding, text generation, coding, and more. 
These evaluations are conducted in both Chinese and English.
In addition, we have specifically evaluated the multilingual performance of instruction-tuned models. 
The results are summarized in Table \ref{tab:eval_v11_EN} for English, Table \ref{tab:eval_v11_ZH} for Chinese, Table \ref{tab:eval_multilingual_70B} for multilingualism of 70B+ Instruct models, and Table \ref{tab:eval_multilingual_14B} for 7B-14B models, respectively.

\paragraph{English \& Chinese Evaluation}

We compare the performance of Qwen2.5-Instruct models against several leading language models, including GPT-4, Claude3.5-sonnet, Qwen2, and Llama-3.1, across both English and Chinese languages. Our analysis focuses on model size and its impact on performance, as well as how our latest Qwen2.5 series compares to previous iterations and competing models.
For smaller models, we observe that the Qwen2.5-0.5B model achieves performance that is on par with or even surpasses the Qwen2-1.5B model. 
This indicates that the Qwen2.5 series has optimized parameter usage, enabling mid-sized models to achieve similar performance levels to larger models from the previous generation.
The Qwen2.5-3B model demonstrates performance that is comparable to the Qwen2-7B model. 
Notably, the Qwen2.5-32B model exhibits a remarkable improvement over the Qwen2-72B model.
Our flagship model, Qwen2.5-72B, further narrows the gap between Qwen and state-of-the-art models like GPT-4 and Claude3.5-sonnet. In particular, Qwen2.5-72B matches or exceeds the performance of Llama-3.1-405B in all metrics except for instruction following. This achievement underscores the competitiveness of Qwen2.5-72B in a wide range of language processing tasks, while also identifying areas for future improvement.
Qwen2.5-Plus addresses the previous shortcomings in Chinese instruction following and further enhances its advantages in other areas.

\paragraph{Multilingual Evaluation}

\begin{table}[t]

\centering

\small

\caption{\textbf{Performance of the 70B+ Instruct models on Multilingual Tasks.}}

\label{tab:eval_multilingual_70B}

\setlength{\tabcolsep}{2.6pt}

\begin{tabular}{@{}lcccccc@{}}

\toprule

\textbf{Datasets} & \textbf{Qwen2-72B} & \textbf{Llama3.1-70B} & \textbf{Qwen2.5-32B} & \textbf{Mistral-Large} & \textbf{GPT4o-mini} & \textbf{Qwen2.5-72B} \\

\midrule

\multicolumn{7}{c}{\textit{Instruction Following}} \\

\midrule

IFEval (multilingual) & 79.69 & 80.47 & 82.68 & 82.69 & 85.03 & \textbf{86.98} \\

\midrule

\multicolumn{7}{c}{\textit{Knowledge}} \\

\midrule

AMMLU (Arabic) & 68.85 & 70.08 & 70.44 & 69.24 & 69.73 & \textbf{72.44} \\

JMMLU (Japanese) & 77.37 & 73.89 & 76.55 & 75.77 & 73.74 & \textbf{80.56} \\

KMMLU (Korean) & 57.04 & 53.23 & 60.75 & 56.42 & 56.77 & \textbf{61.96} \\

IndoMMLU (Indonesian) & 66.31 & 67.50 & 66.42 & 63.21 & 67.75 & \textbf{69.25} \\

TurkishMMLU (Turkish) & 69.22 & 66.89 & 72.41 & 64.78 & 71.19 & \textbf{76.12} \\

okapi MMLU (translated) & 77.84 & 76.49 & 77.16 & 78.37 & 73.44 & \textbf{79.97} \\

\midrule

\multicolumn{7}{c}{\textit{Math Reasoning}} \\

\midrule

MGSM8K (extended) & 82.72 & 73.31 & 87.15 & \textbf{89.01} & 87.36 & 88.16 \\

\midrule

\multicolumn{7}{c}{\textit{Cultural Nuances}} \\

\midrule

BLEnD & 25.90 & 30.49 & 27.88 & 33.47 & \textbf{35.91} & 32.48 \\

\bottomrule

\end{tabular}

\end{table}

\begin{table}[t]

\centering

\caption{\textbf{Performance of the 7B-14B Instruct models on Multilingual Tasks.}}

\label{tab:eval_multilingual_14B}
\small
\setlength{\tabcolsep}{2.6pt}

\begin{tabular}{@{}lccccc@{}}

\toprule

\textbf{Datasets} & \textbf{Qwen2-7B} & \textbf{Llama3.1-8B} & \textbf{Qwen2.5-7B} & \textbf{Gemma2-9B} & \textbf{Qwen2.5-14B} \\

\midrule

\multicolumn{6}{c}{\textit{Instruction Following}} \\

\midrule

IFEval (multilingual) & 51.43 & 60.68 & 74.87 & \textbf{77.47} & 77.08 \\

\midrule

\multicolumn{6}{c}{\textit{Knowledge}} \\

\midrule

AMMLU (Arabic) & 54.87 & 54.28 & 59.78 & 60.26 & \textbf{66.81} \\

JMMLU (Japanese) & 57.71 & 53.26 & 61.88 & 64.59 & \textbf{72.78} \\

KMMLU (Korean) & 43.96 & 42.28 & 46.59 & 46.24 & \textbf{59.71} \\

IndoMMLU (Indonesian) & 54.05 & 53.92 & 56.42 & 61.73 & \textbf{65.09} \\

TurkishMMLU (Turkish) & 49.27 & 45.61 & 54.28 & 55.44 & \textbf{66.85} \\

okapi MMLU (translated) & 60.47 & 55.18 & 66.98 & 46.72 & \textbf{72.12} \\

\midrule

\multicolumn{6}{c}{\textit{Math Reasoning}} \\

\midrule

MGSM8K (extended) & 56.13 & 66.05 & 66.11 & 78.37 & \textbf{82.27} \\

\midrule

\multicolumn{6}{c}{\textit{Cultural Nuances}} \\

\midrule

BLEnD & 22.49 & 19.47 & 23.66 & \textbf{28.31} & 26.99 \\

\bottomrule

\end{tabular}

\end{table}

To comprehensively evaluate the multilingual capabilities of instruction-tuned models, we followed P-MMEval~\citep{pmmeval} and extended several benchmarks as follows:
(1) IFEval (Multilingual): We expanded the IFEval benchmark, originally in English, to include multilingual examples. To ensure language neutrality, we removed instances that contained language-specific content (e.g., "start with letter A"). 
(2) Knowledge Utilization: to assess the knowledge utilization abilities of the Qwen2.5 series models across multiple languages, we employed five MMLU-like benchmarks (multiple-choice format). These benchmarks include: AMMLU (Arabic), JMMLU (Japanese), KMMLU (Korean), IndoMMLU (Indonesian), and TurkishMMLU (Turkish). 
Additionally, we evaluated the models' performance on the translated version of the MMLU benchmark (okapi\_MMLU), which has been adapted into multiple languages from its original English form.
(3) MGSM8K (Extended): Building upon the original MGSM8K benchmark, we extended the language support to include Arabic (ar), Korean (ko), Portuguese (pt), and Vietnamese (vi). 
(4) Cultural Nuances: To evaluate the models' ability to capture cultural nuances, we utilized the BLEnD benchmark~\citep{blend}. This benchmark is specifically designed to test LLMs on their understanding of cultural subtleties. 

Qwen2.5 exhibits competitive performance in instruction following, multilingual knowledge, and mathematical reasoning, aligning well with models of comparable size. Although it shows notable improvements in capturing cultural nuances relative to its predecessor, Qwen2, there remains potential for further refinement in this domain.

\subsubsection{Reward Model}

The reward model serves as the cornerstone for guiding RL processes, and thus we conduct a separate evaluation of the reward model used in the Qwen2.5 series. Our assessment benchmarks encompass Reward Bench~\citep{Lambert2024RewardBenchER}, RMB~\citep{Zhou2024RMBCB}, PPE~\citep{Frick2024HowTE}, and an internally collected out-of-domain Chinese human preference benchmark (Human-Preference-Chinese) to provide a comprehensive analysis.
For comparison, we included baseline models such as Nemotron-4-340B-Reward~\citep{Adler2024Nemotron43T}, Llama-3.1-Nemotron-70B-Reward~\citep{Wang2024HelpSteer2PreferenceCR}, and Athene-RM-70B~\citep{Athene2024}. The results are shown in Table~\ref{tab:eval_reward_bench_metrics}. Overall, our findings indicate that Llama-3.1-Nemotron-70B-Reward excels on the Reward Bench, while Athene-RM-70B performs best on the RMB benchmark. The Qwen2.5-RM-72B, leads in both the PPE and Human-Preference-Chinese evaluations, ranking second only to Athene-RM-70B on the RMB and achieving a performance level comparable to Nemotron-4-340B-Reward on the Reward Bench, albeit slightly behind Llama-3.1-Nemotron-70B-Reward.

Due to the lack of evaluation methods for reward models, current reward models are typically evaluated using Reward Bench. However, our evaluation results from multiple RM benchmarks suggest that over-optimization on a specific benchmark may trigger Goodhart's law~\citep{hoskin1996}, resulting in degraded performance on other benchmarks and potentially impacting downstream alignment performance. This highlights the need for comprehensive evaluation of reward models across diverse benchmarks rather than relying solely on a single benchmark.

More importantly, through iterative experimentation, we have also come to recognize a critical limitation: current reward model evaluation benchmarks do not accurately predict the performance of the RL models trained under their guidance. In other words, a higher score on RM benchmarks does not necessarily correlate with superior performance of the resulting RL model. This insight underscores the need for further research into more predictive evaluation methods for reward models.

\begin{table}[tbp]
\centering
\caption{\textbf{Performance comparison across multiple RM benchmarks.}}
\label{tab:eval_reward_bench_metrics}
\small
\setlength{\tabcolsep}{3pt} %

\begin{tabular}{@{}lcccc@{}}
\toprule
\textbf{Metric} & \textbf{\shortstack{Nemotron-4-340B-\\Reward}} & \textbf{\shortstack{Llama-3.1-Nemotron-\\70B-Reward}} & \textbf{\shortstack{Athene-RM\\-70B}} & \textbf{\shortstack{Qwen2.5-RM\\-72B}} \\
\midrule
\multicolumn{5}{c}{\textit{Reward Bench}} \\
\midrule
Chat & 95.80 & 97.50 & \textbf{98.32} & 97.21 \\
Chat Hard & \textbf{87.10} & 85.70 & 70.61 & 78.73 \\
Safety & 91.50 & \textbf{95.10} & 92.10 & 92.71 \\
Reasoning & 93.60 & \textbf{98.10} & 92.19 & 97.65 \\
Score & 92.00 & \textbf{94.10} & 88.32 & 91.59 \\
\midrule
\multicolumn{5}{c}{\textit{RMB}} \\
\midrule
Helpfulness (BoN) & 48.85 & 61.02 & \textbf{67.24} & 65.72 \\
Helpfulness (Pairwise) & 68.70 & 75.28 & \textbf{80.82} & 78.83 \\
Harmlessness (BoN) & 50.92 & 52.00 & \textbf{67.02} & 56.35 \\
Harmlessness (Pairwise) & 70.84 & 69.96 & \textbf{80.83} & 73.94 \\
Overall & 59.83 & 64.57 & \textbf{73.98} & 68.71 \\
\midrule
\multicolumn{5}{c}{\textit{PPE}} \\
\midrule
Human Preference & 59.28 & 64.32 & \textbf{66.48} & 64.80 \\
IFEval & 62.66 & 63.40 & 62.15 & \textbf{67.97} \\
GPQA & 56.56 & 59.14 & 59.26 & \textbf{59.80} \\
MATH & 65.12 & 69.73 & 79.14 & \textbf{81.48} \\
MBPP-Plus & 49.15 & 55.62 & \textbf{67.97} & 64.34 \\
MMLU-Pro & 69.69 & 70.20 & \textbf{76.95} & 75.66 \\
Objective-Avg & 60.64 & 63.62 & 69.09 & \textbf{69.85} \\
\midrule
\multicolumn{5}{c}{\textit{Human-Preference-Chinese}} \\
\midrule
Accuracy & 50.46 & 59.95 & 61.11 & \textbf{61.27} \\
\bottomrule
\end{tabular}

\end{table}

\begin{table}[tbp]
\centering
\caption{\textbf{Performance of Qwen2.5 Models on RULER.} \textit{YARN+DCA} does not change the model behavior within 32K tokens.}
\label{tab:ruler}
\small
\begin{tabular}{@{}lrlllllll@{}}
\toprule
\multirow{2}{*}{\bf Model} & \bf \multirow{2}[2]{*}{\makecell{\bf Claimed \\ Length}} & \multicolumn{6}{c}{\bf RULER}  \\ 
\cmidrule{3-9}
 & &  \bf Avg.  & \bf 4K   & \bf 8K    & \bf 16K  & \bf 32K  & \bf 64K   & \bf 128K  \\ \midrule
 GLM4-9b-Chat-1M & 1M         & 89.9   & 94.7   & 92.8            & 92.1   & 89.9   & 86.7          & 83.1             \\
Llama-3-8B-Instruct-Gradient-1048k & 1M         & 88.3   & 95.5   & 93.8            & 91.6   & 87.4   & 84.7          & 77.0      \\
Llama-3.1-70B-Instruct & 128K & 89.6 & 96.5 & 95.8 &  95.4 & 94.8  & 88.4 &  66.6 \\
GPT-4o-mini    & 128K       & 87.3   & 95.0   & 92.9            & 92.7   & 90.2   & 87.6          & 65.8                \\ 
GPT-4        & 128K     & 91.6 & 96.6 & 96.3 & 95.2 & 93.2 & 87.0          & 81.2              \\ 
\midrule
\textbf{Qwen2.5-7B-Instruct}  & 128K & 85.4 & 96.7 & 95.1 &93.7 & 89.4&82.3&55.1\\
\textcolor{gray}{\quad w/o DCA + YARN} &  & \textcolor{gray}{80.1} & \textcolor{gray}{96.7} & \textcolor{gray}{95.1} & \textcolor{gray}{93.7} & \textcolor{gray}{89.4} &\textcolor{gray}{74.5}&\textcolor{gray}{31.4} \\
\textbf{Qwen2.5-14B-Instruct} & 128K & 91.4 &97.7 &96.8 &95.9 &93.4 & 86.7 & 78.1 \\
\quad \textcolor{gray}{w/o DCA + YARN} &  & \textcolor{gray}{86.5} & \textcolor{gray}{97.7} & \textcolor{gray}{96.8} & \textcolor{gray}{95.9} & \textcolor{gray}{93.4} & \textcolor{gray}{82.3} & \textcolor{gray}{53.0} \\
\textbf{Qwen2.5-32B-Instruct} & 128K & 92.9 & 96.9 & 97.1 & 95.5 & 95.5 & 90.3 & 82.0  \\
\quad \textcolor{gray}{w/o DCA + YARN} &  & \textcolor{gray}{88.0} & \textcolor{gray}{96.9} & \textcolor{gray}{97.1} & \textcolor{gray}{95.5} & \textcolor{gray}{95.5} & \textcolor{gray}{85.3} & \textcolor{gray}{57.7} \\
\textbf{Qwen2.5-72B-Instruct} & 128K & \bf95.1 & \bf97.7 & \bf 97.2 & \bf 97.7 & \bf 96.5 & \bf 93.0 & \bf 88.4 \\
\quad \textcolor{gray}{w/o DCA + YARN} &  & \textcolor{gray}{90.8} & \textcolor{gray}{97.7} & \textcolor{gray}{97.2} & \textcolor{gray}{97.7} & \textcolor{gray}{96.5} &  \textcolor{gray}{88.5} & \textcolor{gray}{67.0} \\
\midrule
\textbf{Qwen2.5-Turbo}  & 1M & {93.1} & {97.5} & 95.7 & {95.5} & {94.8} & {90.8} & {84.5} \\                 
\bottomrule
\end{tabular}
\end{table}

\subsubsection{Long Context Capabilities}

We utilize three benchmarks to evaluate long context capabilities of Qwen2.5 models: RULER~\citep{hsieh2024ruler}, LV-Eval~\citep{yuan2024lveval}, and Longbench-Chat~\citep{longalign2024bai}. In LV-Eval, we adopt keyword recall as the reported score to mitigate the high rate of false negatives present in the original metrics.

The results are shown in Table \ref{tab:ruler} and Table \ref{tab:lveval}. We can observe that the Qwen2.5 models, after equipping length extrapolation techniques (i.e., DCA + YARN), have demonstrated strong long context processing capabilities on the three datasets. Among them, Qwen2.5-72B-Instruct has shown the strongest performance across all context lengths, significantly outperforming existing open-weight long-context models as well as the proprietary models like GPT-4o-mini and GPT-4.

\begin{table}[tbp]
\caption{\textbf{Performance of Qwen2.5 Models on LV-Eval and LongBench-Chat.} \textit{YARN+DCA} does not change the model behavior within 32k tokens.}
\label{tab:lveval}
\small
\centering
\begin{tabular}{@{}lrrrrrrc@{}}
\toprule
\multirow{2}{*}{\bf Model} & \bf \multirow{2}[2]{*}{\makecell{\bf Claimed \\ Length}} & \multicolumn{5}{c}{\bf LV-Eval} & \multirow{2}[2]{*}{\makecell{\bf LongBench-\\ \bf Chat }} \\ 
\cmidrule{3-7}
 & \bf   & \bf 16k   & \bf 32k    & \bf 64k  & \bf 128k  & \bf 256k    \\ \midrule
GLM4-9B-Chat-1M                    & 1M     &    46.4	&43.2	&42.9	&40.4	&37.0	&7.82                      \\
Llama-3-8B-Instruct-Gradient-1048k & 1M          &   31.7	 &31.8	 & 28.8	 & 26.3	& 21.1 &	6.20            \\
Llama-3.1-70B-Instruct & 128k & 48.6 & 47.4 & 42.9 & 26.2 & N/A & 6.80 \\
GPT-4o-mini                       & 128k    & 52.9	& 48.1	& 46.0	& 40.7	& N/A	& 8.48            \\ 
\midrule
\textbf{Qwen2.5-7B-Instruct}  & 128k & 55.9&49.7&48.0 & 41.1 & 36.9 & 7.42\\
\quad \textcolor{gray}{w/o DCA + YARN}  &  & \textcolor{gray}{55.9}&\textcolor{gray}{49.7}&\textcolor{gray}{33.1}&\textcolor{gray}{13.6}&\textcolor{gray}{0.5} & \textcolor{gray}{-}\\ 
\textbf{Qwen2.5-14B-Instruct} & 128k & 53.0 &50.8 & 46.8 & 43.6 & 39.4 & 8.04  \\
\quad \textcolor{gray}{w/o DCA + YARN}&  & \textcolor{gray}{53.0} &\textcolor{gray}{50.8} & \textcolor{gray}{37.0} & \textcolor{gray}{18.4} & \textcolor{gray}{0.8} & \textcolor{gray}{-}\\
\textbf{Qwen2.5-32B-Instruct} & 128k& 56.0 & 53.6 & 48.8 & 45.3 & 41.0 & 8.70 \\
\quad \textcolor{gray}{w/o DCA + YARN} &  & \textcolor{gray}{56.0} & \textcolor{gray}{53.6} & \textcolor{gray}{40.1} & \textcolor{gray}{20.5} & \textcolor{gray}{0.7} & \textcolor{gray}{-}\\
\textbf{Qwen2.5-72B-Instruct} & 128k & \bf60.4 & \bf 57.5 &  \bf53.9 & \bf50.9 & \bf45.2 & \bf 8.72 \\
\quad \textcolor{gray}{w/o DCA + YARN} &  & \textcolor{gray}{60.4} & \textcolor{gray}{57.5} & \textcolor{gray}{47.4} & \textcolor{gray}{27.0} & \textcolor{gray}{2.4} & \textcolor{gray}{-} \\
\midrule
\textbf{Qwen2.5-Turbo}             & 1M     & 53.4	& 50.0	&45.4 	& 43.9	& 38.0&	8.34        \\   
\bottomrule
\end{tabular}
\end{table}

\begin{figure}[tbp]
    \centering
    \includegraphics[width=\textwidth]{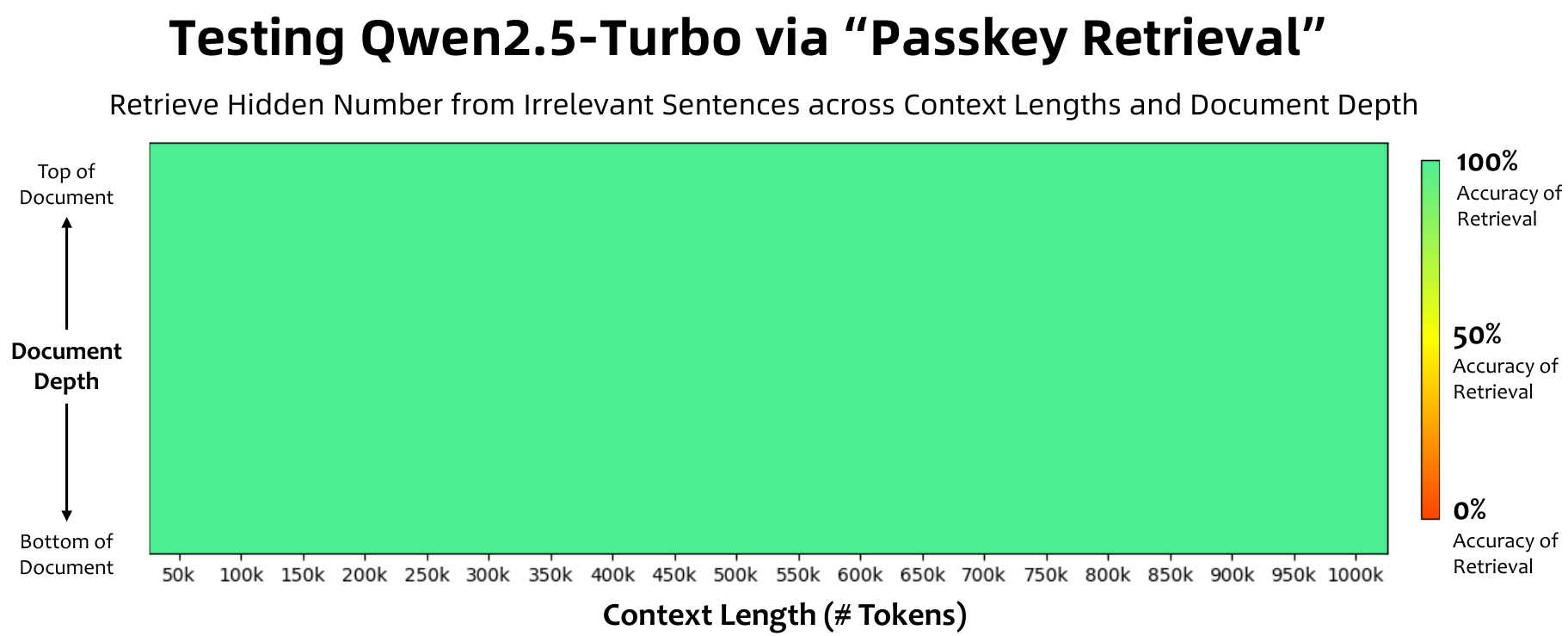}
    \caption{\textbf{Performance of Qwen2.5-Turbo on Passkey Retrieval Task with 1M Token Lengths.}}
    \label{fig:passkey_retrieval}
\end{figure}

\begin{figure}[tb]
    \centering
    \includegraphics[width=0.8\textwidth]{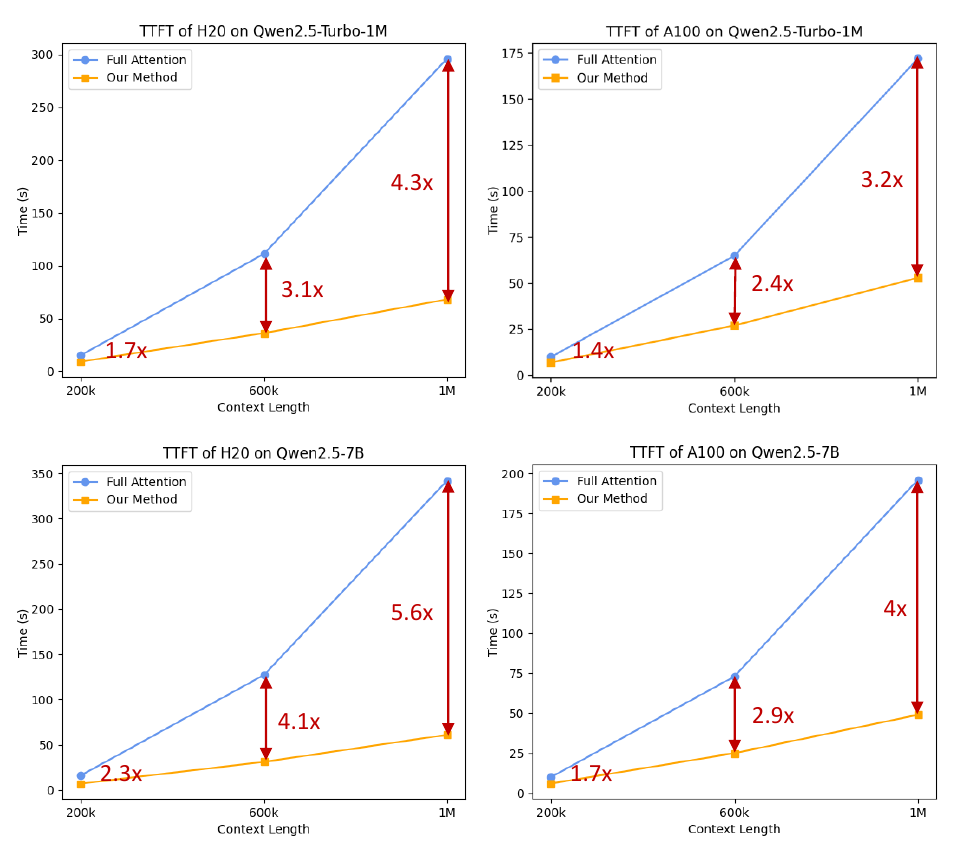}
    \caption{\textbf{TTFT (Time To First Token) of Qwen2.5-Turbo and Qwen2.5-7B with Full Attention and Our Method.}}
    \label{fig:inference_time}
\end{figure}

Furthermore, as shown in Figure \ref{fig:passkey_retrieval}, Qwen2.5-Turbo achieves 100\% accuracy in the 1M-token passkey retrieval task, demonstrating its exceptional ability to capture detailed information from ultra-long contexts. 
We develop a sparse attention mechanism based on Minference~\citep{jiang2024minference} to significantly enhance inference speed, which is critical for user experience when processing long contexts. For sequences of 1M tokens, this approach reduces the computational load of the attention mechanism by 12.5 times. Figure \ref{fig:inference_time} illustrates the time to first token (TTFT) of Qwen2.5-Turbo across various hardware configurations, where our method achieves a 3.2 to 4.3 times speedup.

\section{Conclusion}
\label{sec:conclusion}

Qwen2.5 represents a significant advancement in large language models (LLMs), with enhanced pre-training on 18 trillion tokens and sophisticated post-training techniques, including supervised fine-tuning and multi-stage reinforcement learning. These improvements boost human preference alignment, long text generation, and structural data analysis, making Qwen2.5 highly effective for instruction-following tasks.
Available in various configurations, Qwen2.5 offers both open-weight from 0.5B to 72B parameters and proprietary models including cost-effective MoE variants like Qwen2.5-Turbo and Qwen2.5-Plus. Empirical evaluations show that Qwen2.5-72B-Instruct matches the performance of the state-of-the-art Llama-3-405B-Instruct, despite being six times smaller. Qwen2.5 also serves as a foundation for specialized models, demonstrating its versatility for domain-specific applications.
We believe that Qwen2.5's robust performance, flexible architecture, and broad availability make it a valuable resource for both academic research and industrial applications, positioning it as a key player of future innovations.

In the future, we will focus on advancing robust foundational models. 
First, we will iteratively refine both base and instruction-tuned large language models (LLMs) by incorporating broader, more diverse, higher-quality data.
Second, we will also continue to develop multimodal models. Our goal is to integrate various modalities into a unified framework. This will facilitate seamless, end-to-end information processing across textual, visual, and auditory domains.
Third, we are committed to enhancing the reasoning capabilities of our models. This will be achieved through strategic scaling of inference compute resources. 
These efforts aim to push the boundaries of current technological limitations and contribute to the broader field of artificial intelligence.

\section{Authors}

\textbf{Core Contributors:} An Yang, Baosong Yang, Beichen Zhang, Binyuan Hui, Bo Zheng, Bowen Yu, Chengyuan Li, Dayiheng Liu, Fei Huang, Haoran Wei, Huan Lin, Jian Yang, Jianhong Tu, Jianwei Zhang, Jianxin Yang, Jiaxi Yang, Jingren Zhou, Junyang Lin, Kai Dang, Keming Lu, Keqin Bao, Kexin Yang, Le Yu, Mei Li, Mingfeng Xue, Pei Zhang, Qin Zhu, Rui Men, Runji Lin, Tianhao Li, Tianyi Tang, Tingyu Xia, Xingzhang Ren, Xuancheng Ren, Yang Fan, Yang Su, Yichang Zhang, Yu Wan, Yuqiong Liu, Zeyu Cui, Zhenru Zhang, Zihan Qiu

\textbf{Contributors:} Biao Sun, Bin Luo, Bin Zhang, Binghai Wang, Chaojie Yang, Chang Si, Cheng Chen, Chengpeng Li, Chujie Zheng, Fan Hong, Guanting Dong, Guobin Zhao, Hangrui Hu, Hanyu Zhao, Hao Lin, Hao Xiang, Haoyan Huang, Humen Zhong, Jialin Wang, Jialong Tang, Jiandong Jiang, Jianqiang Wan, Jianxin Ma, Jianyuan Zeng, Jie Zhang, Jin Xu, Jinkai Wang, Jinzheng He, Jun Tang, Ke Yi, Keqin Chen, Langshi Chen, Le Jiang, Lei Zhang, Liang Chen, Man Yuan, Mingkun Yang, Minmin Sun, Na Ni, Nuo Chen, Peng Wang, Peng Zhu, Pengcheng Zhang, Pengfei Wang, Qiaoyu Tang, Qing Fu, Rong Zhang, Ru Peng, Ruize Gao, Shanghaoran Quan, Shen Huang, Shuai Bai, Shuang Luo, Sibo Song, Song Chen, Tao He, Ting He, Wei Ding, Wei Liao, Weijia Xu, Wenbin Ge, Wenbiao Yin, Wenyuan Yu, Xianyan Jia, Xianzhong Shi, Xiaodong Deng, Xiaoming Huang, Ximing Zhou, Xinyu Wang, Xipin Wei, Xuejing Liu, Yang Liu, Yang Yao, Yang Zhang, Yibo Miao, Yidan Zhang, Yikai Zhu, Yinger Zhang, Yong Jiang, Yong Li, Yongan Yue, Yuanzhi Zhu, Yunfei Chu, Zekun Wang, Zhaohai Li, Zheren Fu, Zhi Li, Zhibo Yang, Zhifang Guo, Zhipeng Zhang, Zhiying Xu, Zile Qiao, Ziye Meng

\bibliography{biblio}
\bibliographystyle{colm2024_conference}

\end{document}